\documentclass[pdflatex,sn-mathphys-num]{sn-jnl}

\usepackage{graphicx}%
\usepackage{multirow}%
\usepackage{amsmath,amssymb,amsfonts}%
\usepackage{amsthm}%
\usepackage{mathrsfs}%
\usepackage[title]{appendix}%
\usepackage{textcomp}%
\usepackage{manyfoot}%
\usepackage{booktabs}%
\usepackage{algorithm}%
\usepackage{algorithmicx}%
\usepackage{algpseudocode}%
\usepackage{listings}%
\usepackage{subcaption}
\usepackage[normalem]{ulem}

\usepackage{mathtools}
\usepackage{amsthm}
\usepackage{pifont}
\usepackage{multirow}
\usepackage{makecell} 
\usepackage{adjustbox}

\usepackage{booktabs}
\usepackage[table,xcdraw]{xcolor}
\usepackage{makecell}
\usepackage{array}
\usepackage{multirow}
\usepackage[normalem]{ulem}
\usepackage[numbers]{natbib}

\usepackage[capitalize]{cleveref}
\crefname{section}{Sec.}{Secs.}
\Crefname{table}{Table.}{Tables.}
\crefname{figure}{Fig.}{Figs.}

\newcommand{\cmark}{\ding{51}} 
\newcommand{\xmark}{\ding{55}} 
\newcommand{\parahead}[1]{\noindent\textbf{#1.} \enskip}
\newcommand{\quotes}[1]{\textbf{\textit{``#1''}}}

\newcommand{\changes}[1]{\textcolor{black}{#1}}
\newcommand{\edits}[1]{\textcolor{black}{#1}}

\newcommand{\icare}{\textcolor{black}{\textbf{ICARE}}}
\newcommand{\icarefull}{\textcolor{black}{\textbf{I}nterpretable and \textbf{C}linically-grounded \textbf{A}gent-based \textbf{R}eport \textbf{E}valuation}}

\newcommand{\icarefullnobold}{\textcolor{black}{Interpretable and Clinically-grounded Agent-based Report Evaluation}}

\theoremstyle{thmstyleone}%
\theoremstyle{thmstyletwo}%

\theoremstyle{thmstylethree}%

\makeatletter
\patchcmd{\@maketitle}{\vskip20pt}{\vskip2pt}{}{}
\makeatother
\begin{document}
\title[Clinically Grounded Agent-based Report Evaluation: An Interpretable Metric for Radiology Report Generation
]{Clinically Grounded Agent-based Report Evaluation: An Interpretable Metric for Radiology Report Generation
}
\author*[1,2]{\fnm{Radhika} \sur{Dua}}\email{rd3571@nyu.edu}

\author[4]{\fnm{Young Joon (Fred)} \sur{Kwon}}\email{YoungJoon.Kwon@nyulangone.org}

\author[4]{\fnm{Siddhant} \sur{Dogra}}\email{Siddhant.Dogra@nyulangone.org}
\employmentnote{Siddhant is a part-time employee of a2z Radiology AI and holds stock equity in the company.}
\equalcont{These authors contributed equally to this work.}

\author[4]{\fnm{Daniel} \sur{Freedman}}\email{Daniel.Freedman@nyulangone.org}
\equalcont{These authors contributed equally to this work.}

\author[4]{\fnm{Diana} \sur{Ruan}}\email{Diana.Ruan@nyulangone.org}
\equalcont{These authors contributed equally to this work.}

\author[4]{\fnm{Motaz} \sur{Nashawaty}}\email{Motaz.Nashawaty@nyulangone.org}
\equalcont{These authors contributed equally to this work.}

\author[4]{\fnm{Danielle} \sur{Rigau}}\email{Danielle.Rigau@nyulangone.org}
\equalcont{These authors contributed equally to this work.}

\author[2,5]{\fnm{Daniel Alexander} \sur{Alber}}\email{Daniel.Alber@nyulangone.org}
\equalcont{These authors contributed equally to this work.}

\author[6,7,8]{\fnm{Kang} \sur{Zhang}}\email{kang.zhang@gmail.com}

\author[1,3]{\fnm{Kyunghyun} \sur{Cho}}\email{kc119@nyu.edu}

\author[1,2,4]{\fnm{Eric Karl} \sur{Oermann}}\email{eric.oermann@nyulangone.org}

\affil*[1]{\orgdiv{Center for Data Science}, \orgname{New York University}, \orgaddress{\street{60 5th Ave}, \city{New York}, \postcode{100190}, \state{NY}, \country{USA}}}

\affil[2]{\orgdiv{Department of Neurosurgery}, \orgname{NYU Langone Health}, \orgaddress{\street{450 First Avenue}, \city{New York City}, \postcode{10019}, \state{NY}, \country{USA}}}

\affil[3]{\orgdiv{Prescient Design}, \orgname{Genentech}, \orgaddress{\street{149 5th Ave. 3rd floor}, \city{New York}, \postcode{10019}, \state{NY}, \country{USA}}}

\affil[4]{\orgdiv{Department of Radiology}, \orgname{NYU Langone Health}, \orgaddress{\street{450 First Avenue}, \city{New York City}, \postcode{10019}, \state{NY}, \country{USA}}}

\affil[5]{\orgdiv{NYU Grossman School of Medicine}, \orgname{NYU Langone Health}, \orgaddress{\street{450 First Avenue}, \city{New York City}, \postcode{10019}, \state{NY}, \country{USA}}}

\affil[6]{\orgdiv{National Clinical Eye Research Center, Eye Hospital}, \orgname{Wenzhou Medical University}, \orgaddress{\city{Wenzhou}, \postcode{325000}, \country{China}}}

\affil[7]{\orgdiv{Institute for Clinical Data Science}, \orgname{Wenzhou Medical University}, \orgaddress{\city{Wenzhou}, \postcode{325000}, \country{China}}}

\affil[8]{\orgdiv{Institute for AI in Medicine and Faculty of Medicine}, \orgname{Macau University of Science and Technology}, \orgaddress{\city{Macau}, \postcode{999078}, \country{China}}}


\abstract{
\begin{abstract}

Radiological imaging is central to diagnosis, treatment planning, and clinical decision-making. Vision-language foundation models have spurred interest in automated radiology report generation (RRG), but safe deployment requires reliable clinical evaluation of generated reports. Existing metrics often rely on surface-level similarity and/or behave as black boxes, lacking interpretability. We introduce \icare(\icarefull), an interpretable evaluation framework leveraging large language model agents and dynamic multiple-choice question answering (MCQA). Two agents, each with either the ground-truth or generated report, generate clinically meaningful questions and quiz each other. Agreement on answers captures preservation and consistency of findings, serving as interpretable proxies for clinical precision and recall. By linking scores to question–answer pairs, \icare\ enables transparent, and interpretable assessment. Clinician studies show \icare\ aligns significantly more with expert judgment than prior metrics. Perturbation analyses confirm sensitivity to clinical content and reproducibility, while model comparisons reveal interpretable error patterns.

\end{abstract}

}
\maketitle

Radiology reports are essential for accurate diagnosis, treatment planning, and communication among clinical teams. Traditionally, these reports are written by radiologists based on their interpretation of imaging studies such as chest X-rays, CT scans, or MRIs. However, this process is time-consuming, cognitively demanding, and requires significant clinical expertise. With the increasing volume of imaging studies and a global shortage of radiologists, many healthcare systems are facing mounting pressure on clinical workflows. This strain often results in reporting delays and increases the risk of diagnostic errors.
In response to these challenges, automated radiology report generation (RRG) systems have emerged as a promising solution. Recent methods span a range of modeling approaches, including vision-language models like Flamingo-CXR \cite{Tanno2023ConsensusDA}, CNN-LSTM architectures with attention mechanisms \cite{Sirshar2022AttentionBA}, and knowledge-enhanced models incorporating structured medical information \cite{Kale2023ReplaceAR}. More recent advances leverage large multimodal models tailored for radiology, such as MAIRA-1\cite{Hyland2023MAIRA1AS}, MAIRA-2\cite{Bannur2024MAIRA2GR}, LLAVA-Rad\cite{Chaves2024ACA}, and Radialog\cite{Pellegrini2023RaDialogAL}, which integrate domain-specific vision encoders with LLMs to improve clinical accuracy.
Other approaches like MedPaLM-M\cite{Singhal2022LargeLM} and MedVersa\cite{Zhou2024MedVersaAG} highlight the growing focus on scaling, instruction tuning, and factuality evaluation in medical report generation. These models aim to reduce radiologists’ workload, improve the consistency and clarity of reports, and enhance the scalability of radiological services.

\begin{figure}[htbp]
    \centering
    \includegraphics[width=\linewidth]{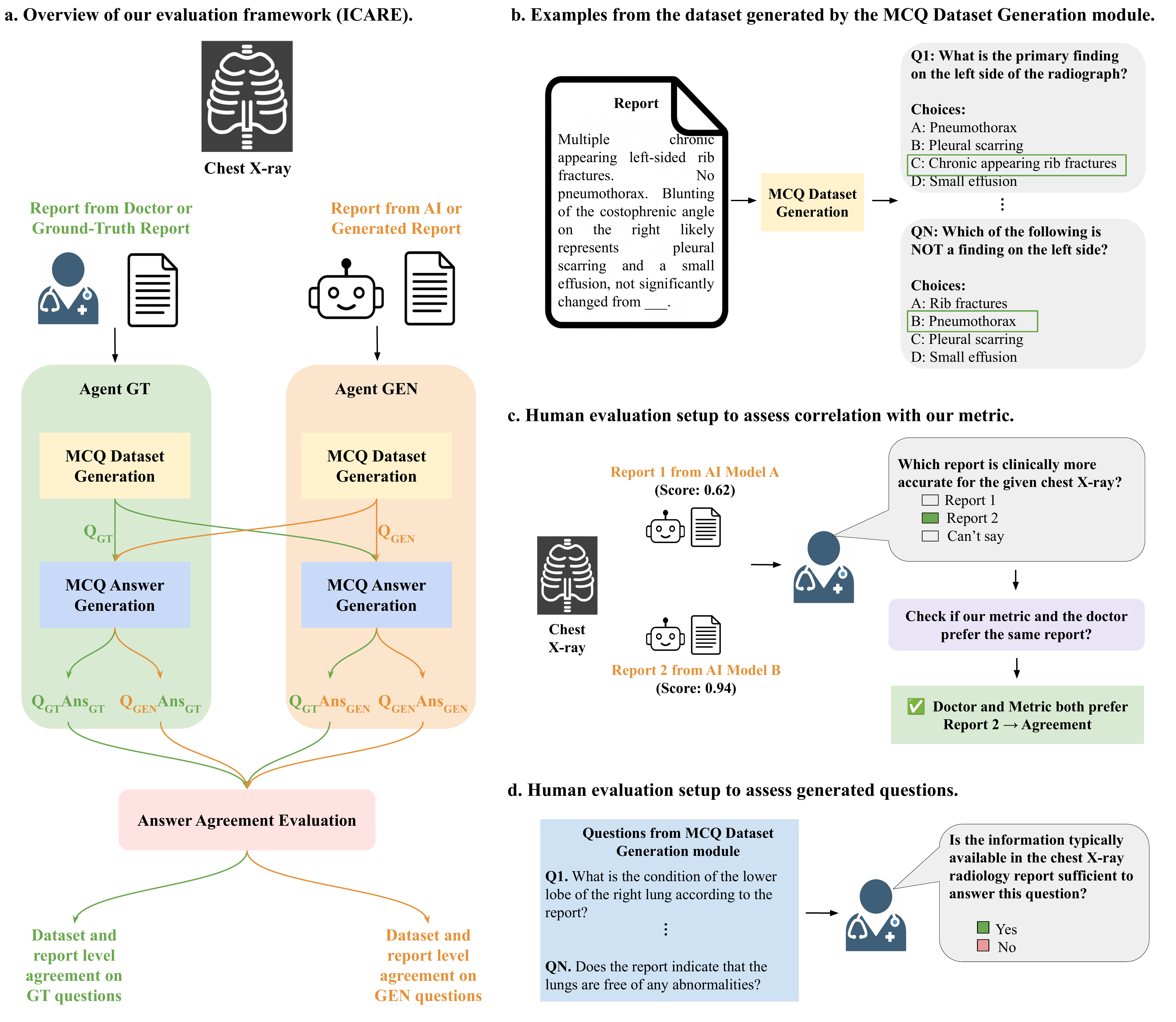}
    \caption{\textbf{Overview of our evaluation framework and human validation process.}
    \textbf{(a)} \icare: \icarefull. Two report-aware agents, AgentGT (ground truth) and AgentGEN (generated), independently generate and answer clinically meaningful multiple-choice questions based solely on their respective input reports. The resulting answers are compared through an external agreement module to assess clinical similarity. Agreement on ground-truth questions estimates precision(\textbf{\icare-GT}), while agreement on generated-report questions estimates recall(\textbf{\icare-GEN}).
    \textbf{(b)} Examples of MCQs generated by the framework, capturing diverse findings such as pleural effusion, rib fractures, and cardiomegaly.
    \textbf{(c)} Human evaluation setup to assess alignment between our metric and expert judgment. Clinicians are shown a chest X-ray and two corresponding reports, and are asked to select the more clinically accurate report or choose ``Can’t say.''
    \textbf{(d)} Human evaluation of question quality. Clinicians assess whether each question can be answered using only the information typically available in the chest X-ray report.}
    \label{fig:panel1}
\end{figure}

Before such systems can be safely deployed in clinical settings to assist radiologists, it is essential to rigorously evaluate whether the generated reports are comparable to those written by experts, both in terms of clinical content and reliability. This raises a fundamental question: \quotes{what are the essential criteria that an evaluation metric must satisfy to be clinically useful?} We argue that three properties are necessary. First, \textbf{semantic understanding}: the metric should assess whether the reports convey the same clinical information, rather than matching surface-level text. It must capture medically relevant content, regardless of phrasing. Second, \textbf{interpretability}: the metric should provide a clear and direct connection between its score and the clinical content being evaluated. It should allow both clinicians and model developers to examine which specific elements of the reports align or diverge, providing transparency into how the score was derived. This level of insight is critical for evaluation in high-stakes settings such as healthcare. Third, \textbf{scalability}: the metric should be able to evaluate large volumes of reports efficiently, while preserving both semantic understanding and interpretability.

Although medical experts can reliably assess the quality of generated reports, manual evaluation is time-consuming, subjective, and difficult to scale. In real-world settings, where hundreds or thousands of reports must be evaluated to monitor model performance or guide clinical deployment, manual review is impractical. This has led to growing interest in automated evaluation metrics. Despite significant progress, most existing metrics fall short of these properties. While they often meet the demands of scalability, they do not fully satisfy the requirements for semantic understanding nor interpretability. Traditional string-matching metrics such as BLEU\citep{Papineni2002BleuAM} and ROUGE\citep{Lin2004ROUGEAP}, adapted from natural language processing, rely on surface-level word overlap and often fail 
to capture clinical equivalence between reports.
Embedding-based metrics such as BERTScore\citep{Zhang2019BERTScoreET} offer improved semantic matching but behave as black boxes, providing limited insights into which specific clinical findings contribute to the score. 
Domain-specific metrics such as F1-CheXpert\citep{Irvin2019CheXpertAL}, SembScore \citep{Smit2020CheXbertCA}, and F1-RadGraph\citep{Yu2022EvaluatingPI} compare reports using structured clinical labels, vectors, or entities, but still lack transparency into what specific differences influence the final score. RadCliQ\citep{Yu2022EvaluatingPI} combines these metrics via regression to approximate radiologist preferences, but lacks clarity on which aspects influence the score. More recent approaches leverage large language models: GREEN\citep{Ostmeier2024GREENGR} distills GPT-based similarity judgments; FineRadScore\citep{Huang2024FineRadScoreAR} performs line-level corrections with severity annotations; RaTEScore\citep{Zhao2024RaTEScoreAM} matches named entities using clinically-aware embeddings; G-Rad\citep{Chaves2024TrainingSM} learns a report ranking model from expert preferences; \edits{and RadFact\citep{Bannur2024MAIRA2GR} assesses factual consistency by checking whether each sentence in one report is supported by the full text of the other using large language model-based inference.} Despite these advancements, most metrics still compromise a subset of interpretability, semantic depth, and scalability. Our MCQA-based evaluation addresses this gap by linking answer agreement to clinically meaningful question–answer pairs.

To this end, we propose \icare(\icarefull), a clinically grounded evaluation framework for radiology report generation. Our approach conceptualizes two report-aware agents operating in parallel: one agent receives the ground-truth report as input, and the other receives the generated report. Each agent first generates a set of multiple-choice questions based on its respective input report using a large language model. A filtering step removes generic or broadly answerable questions, ensuring that the remaining questions require report-specific information to answer correctly.
Each agent then answers all available questions: both the questions produced from the ground-truth report and those produced from the generated report. Specifically, the ground-truth agent answers both sets using the ground-truth report, and the generated agent answers both sets using the generated report. After answering, we compare the outputs of the two agents for each question. Agreement on questions originating from the ground-truth report measures whether the generated report preserves clinically important information, interpreted as a measure of \textbf{precision}. Agreement on questions originating from the generated report assesses whether any additional content introduced is clinically consistent with the ground-truth, interpreted as a measure of \textbf{recall}. The overall similarity score summarizes the degree of clinical alignment between the two reports based on answer consistency.

Our evaluation framework satisfies all three criteria necessary for clinical use. It captures \textbf{semantic understanding} by determining whether two reports lead to consistent clinical conclusions through question answering, even when their phrasing differs. For example, if one report describes ``bilateral infiltrates'' and another mentions ``diffuse opacities in both lungs,'' both should yield the same answer to a question about bilateral involvement. The method is \textbf{interpretable} because each similarity score is directly linked to specific question–answer pairs, allowing clinicians and model developers to inspect which clinical elements contributed to agreement or disagreement. Finally, it is \textbf{scalable}, as the entire process of question generation, answering, and evaluation is automated and can be applied across large report datasets without requiring human review.

To validate our approach, we perform human studies to assess the clinical relevance of the generated questions and to measure the alignment of our \icare and prior metrics with expert judgment. We conduct perturbation experiments to evaluate the sensitivity of our metric to variations in report content, and stability analyses to assess consistency across different runs of question generation and answering. We apply our framework across multiple radiology report generation models to systematically evaluate their performance with respect to \icare, which captures the core clinical criteria of semantic understanding, interpretability, and scalability. Finally, we analyze the agreement patterns to gain interpretable insights into the quality of generated reports, identifying which types of clinical information are preserved, omitted, or altered relative to the ground-truth reports.

To summarize, our contributions are:


\begin{itemize}

    \item \textbf{\icarefullnobold:} \edits{We introduce a clinically grounded, agent-based evaluation framework for radiology report generation. Each agent, assigned either the ground-truth or the generated report, generates clinically meaningful multiple-choice questions based on its report. Both agents are then quizzed on both sets of questions, and their answers are compared to assess the alignment of clinical content between the reports. Our framework yields two agreement-based scores that act as interpretable proxies for clinical precision and recall. These scores quantify the preservation of key findings and the consistency of additional content, offering fine-grained insight into model behavior. Each score is directly grounded in question--answer agreement, allowing clinicians and developers to trace evaluation results back to specific findings and questions, enhancing transparency. The evaluation is fully automated, scalable to large datasets, and generalizable to other imaging modalities and clinical text generation tasks beyond radiology.}

    \item \edits{\textbf{Validation through expert studies and error analysis:} We conduct human evaluations with board-certified clinicians. The first study confirms that asking such multiple-choice questions in the context of chest X-rays is clinically appropriate, supporting the relevance of our question generation approach. A second study shows that \icare\ aligns more closely with expert preferences than prior metrics, while capturing clinically meaningful differences in report quality. To further interpret model behavior, we semantically cluster clinical questions and identify systematic error patterns such as omission and hallucination, offering insight into model limitations across clinical concepts}

    \item \textbf{Robustness and sensitivity:} \edits{We show that ICARE is sensitive to clinically meaningful perturbations, with scores degrading predictably as report content is distorted. At the same time, it exhibits strong stability across different runs of question and answer generation, demonstrating its reliability and reproducibility for real-world use.}

\end{itemize}

\section{Results}\label{results}
\subsection{A clinically grounded, interpretable framework for evaluating radiology report generation}

\begin{figure}[htbp]
    \centering
    \includegraphics[width=\linewidth]{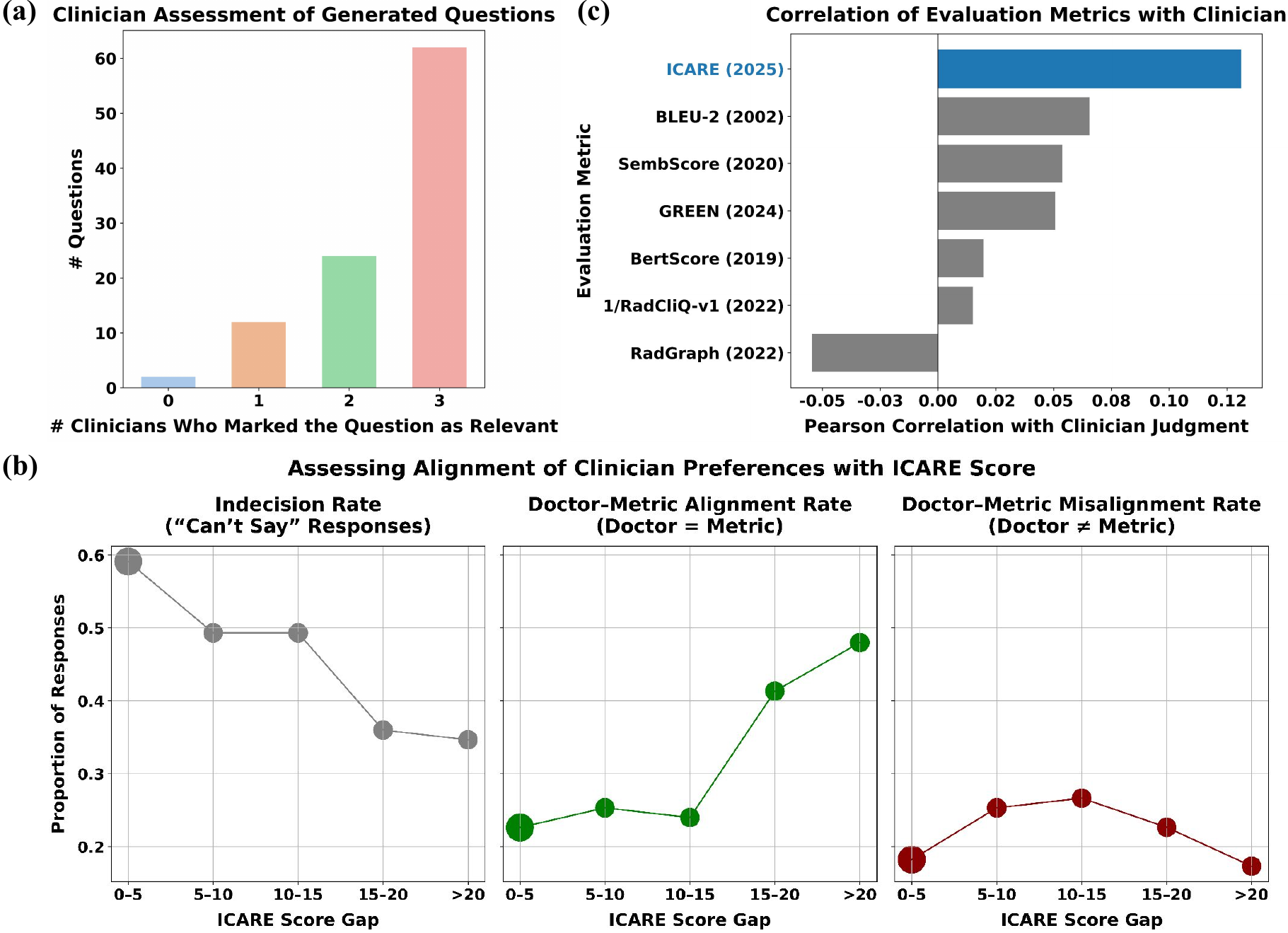}
    \caption{
    \small{
    \textbf{Human validation of question relevance and alignment of evaluation scores with expert preference.}
    \textbf{(a)} Radiologist assessment of whether each multiple-choice question can be answered using only the chest X-ray report. Most questions received agreement from at least two of three clinicians, confirming their suitability for evaluating report content.
    \textbf{(b)} Clinician preferences across report pairs, shown in three plots: Indecision Rate (left), Alignment Rate (center), and Misalignment Rate (right), grouped by \icare\ score gap. When the score gap is small, clinicians often selected ``Can't say,'' indicating uncertainty between similarly scored reports. As the gap increases, ``Can't say'' responses decrease and alignment rises, showing that both clinicians and the metric are more confident in distinguishing report quality. Misalignment remains low throughout. These trends highlight that the metric aligns with expert judgment and reflects meaningful clinical differences.  Dot size reflects the number of samples in each \icare\ score gap bin.
    \textbf{(c)} Correlation between preferences based on clinician judgement and different automatic evaluation metrics. Our metric \icare\ shows the strongest correlation, indicating that it most closely captures clinician judgments of report quality across samples.}
    }
    \label{fig:panel2}
\end{figure}

Our evaluation framework, \icare, uses two independent report-aware agents to assess the clinical similarity between a generated radiology report and its corresponding ground truth report. As shown in \cref{fig:panel1}, one agent operates on the ground truth report and the other on the generated report. Each agent performs two key tasks: it first generates multiple-choice questions based solely on its assigned report (MCQ Dataset Generation), and then answers both its own and the other agent's questions using only its input report (MCQ Answer Generation). 
We use the LLAMA 3.1 70B language model\footnote{While we use LLAMA 3.1 70B in our implementation, our framework(\icare) isn't specific to any particular language model and can be applied with any sufficiently capable language model.}~\citep{Dubey2024TheL3} and task-specific prompts for both multiple-choice question generation and answering, as illustrated in Extended Dataset \cref{fig:prompts}.

To ensure that the questions are clinically meaningful and specific to the report, we apply a filtering step that removes any question which can be answered correctly without access to the report. In addition, we apply answer choice shuffling before and after filtering to mitigate positional biases in the placement of correct options. This results in a balanced and clinically grounded set of questions for evaluation.

Agreement on questions produced from the ground truth report is denoted as \textbf{\icare-GT} and reflects whether the generated report preserves key clinical information, serving as a proxy for precision. Agreement on questions produced from the generated report is denoted as \textbf{\icare-GEN} and captures whether any new content introduced remains consistent with the ground truth, acting as a proxy for recall. We also compute an overall score, \textbf{\icare-AVG}, by averaging \icare-GT and \icare-GEN.

\textbf{We compute these agreement scores at two levels:} at the \textit{dataset level}, by averaging agreement scores across all report pairs, and at the \textit{report level}, by calculating agreement for each individual report. This enables both high-level benchmarking and fine-grained analysis of model outputs.

\edits{We apply this framework to evaluate three pretrained radiology report generation models on the IU X-ray\citep{DemnerFushman2015PreparingAC}\ dataset: CheXpertPlus trained on MIMIC\citep{CheXpertPH}\footnote{https://huggingface.co/IAMJB/mimic-cxr-findings-baseline}, CheXpertPlus trained on CheXpertPlus and MIMIC\citep{CheXpertPH},\footnote{https://huggingface.co/IAMJB/chexpert-mimic-cxr-findings-baseline} and MAIRA2\citep{Bannur2024MAIRA2GR}.\footnote{https://huggingface.co/microsoft/maira-2} To ensure fair comparison, all prior metrics were re-computed. The metrics BLEU-2, BERTScore, SembScore, RadGraph, and 1/RadCliqQ-v1 were evaluated using a recent consolidated codebase,\footnote{https://github.com/rajpurkarlab/CXR-Report-Metric} while GREEN was computed using its official repository.\footnote{https://github.com/ostmeier/green}}
\changes{Although our method is broadly applicable across imaging modalities and report generation tasks, we evaluate it on chest X-rays due to their clinical importance in diagnosing cardiopulmonary conditions and their widespread availability in both public and private datasets.} Further implementation details, including dataset construction, prompt design, filtering criteria, and evaluation setup, are provided in the Method section.

\subsection{Clinicians validate the quality of MCQA-generated questions}

To evaluate the clinical relevance of the questions used in our framework, we conducted a study with six clinicians. 
As illustrated in Figure~\ref{fig:panel1}(d), each clinician was presented with a multiple-choice question from our MCQA dataset and asked to assess whether it would be appropriate to ask in the context of chest radiography, and whether it could be answered using only the information typically available in a chest X-ray report.
We sampled 100 multiple-choice questions from our MCQA dataset, covering a broad range of clinical concepts listed in \cref{fig:cluster-analysis-overall}~(d).  Each question was independently evaluated by three clinicians, resulting in a total of 300 assessments.

Figure~\ref{fig:panel2}~(a) shows how many clinicians endorsed each question as appropriate, grouping questions by the number of endorsing raters (0, 1, 2, or 3). Most questions were endorsed by all three raters, demonstrating strong consensus on their clinical appropriateness. Across 300 total responses, 246 judged the questions as appropriate and answerable using only the chest X-ray report, yielding a high overall endorsement rate (82.0\%). These results confirm that the questions used in our framework are well-suited for evaluating the content quality of radiology reports.

We further analyzed the questions by clinical category to identify which types are most appropriate to ask of chest X-ray reports. As shown in Extended Dataset~\cref{fig:human_eval_ques_with_clusters}, questions related to commonly observed and radiographically accessible findings, such as pleural effusion, lung opacity, and heart size, received the highest levels of endorsement. In contrast, clusters involving less routinely documented or more subtle findings, including thoracic spine changes and calcified granulomas, received lower agreement. This analysis highlights which categories of clinical content align best with the scope of chest radiograph interpretation and supports the use of these questions for evaluating report quality.

\subsection{Alignment of Agreement Scores with Human Judgment}

\parahead{Clinician study design}To assess the clinical validity of our MCQA-based agreement scores, we conducted a human evaluation study involving six board-certified clinicians. We evaluated 154 samples, with each sample independently reviewed by three clinicians. As shown in \cref{fig:panel1}~(c), each clinician was presented with a chest X-ray image and a pair of corresponding reports generated by AI models. They were instructed to select the report that better described the chest X-ray, focusing exclusively on clinical content rather than style or formatting. Clinicians were asked to choose one report if it clearly conveyed the findings more accurately, or select ``Can't say'' if both reports were equally good or poor, or if the difference was too small to confidently choose one over the other.

\parahead{Does our metric align with clinician judgments?}~\cref{fig:panel2}~(b) presents clinician preferences across report pairs, grouped by \icare\ score gap (i.e., the difference in \icare-AVG between the two reports), using three plots: Indecision Rate (left), Alignment Rate (center), and Misalignment Rate (right). Each point in the plots corresponds to a bin of report pairs, and the size of the point reflects the number of samples within that bin. The Indecision Rate plot shows that when the \icare\ score gap is small, clinicians frequently selected ``Can't say,'' suggesting uncertainty between similarly scored reports. As the score gap increases, ``Can't say'' responses decrease, indicating that both the metric and clinicians find clearer differences between reports. In parallel, the Alignment Rate increases with score gap, showing that clinicians consistently favor the report with the higher \icare\ score when the difference is more pronounced. Meanwhile, the Misalignment Rate remains low across all gap bins, reflecting few strong disagreements between clinician judgment and the metric. Together, these trends demonstrate that the metric reliably captures differences in report quality that matter to clinicians and aligns well with expert judgment across varying levels of report similarity.

\parahead{Does our metric align better than prior metrics?} 
To more directly compare our metric with prior evaluation methods, we analyzed how well each metric's preferences align with individual clinician responses. For each doctor response on a report pair, we assigned a label of +1 if the doctor preferred Report 1, -1 if they preferred Report 2, and 0 if they selected ``Can't say.'' Similarly, for each metric, we assigned a label of +1 if Report 1 received a higher score, -1 if Report 2 did, and 0 if the scores were equal. These labels are ordinal, capturing both agreement and the severity of disagreement. A perfect match occurs when the human and metric labels are identical. A full mismatch (e.g., +1 vs. -1) reflects strong disagreement, while a partial mismatch (e.g., +1 vs. 0) reflects a weaker conflict where one party is undecided. We computed Pearson correlation between these label vectors across 459 individual responses. As shown in ~\cref{fig:panel2}(c), \icare\ achieves the highest correlation with expert preferences. Despite the proliferation of recent evaluation metrics, most exhibit weak or negative alignment with clinicians, underscoring the need for clinically grounded assessment of evaluation metrics themselves. These findings reinforce that our metric not only provides interpretable clinical signals but also most closely reflects how radiologists judge report quality.

\subsection{Quantitative evaluation of our metric on radiology report generation models}


\begin{figure}[ht!]
    \centering
    \includegraphics[width=0.92\columnwidth]{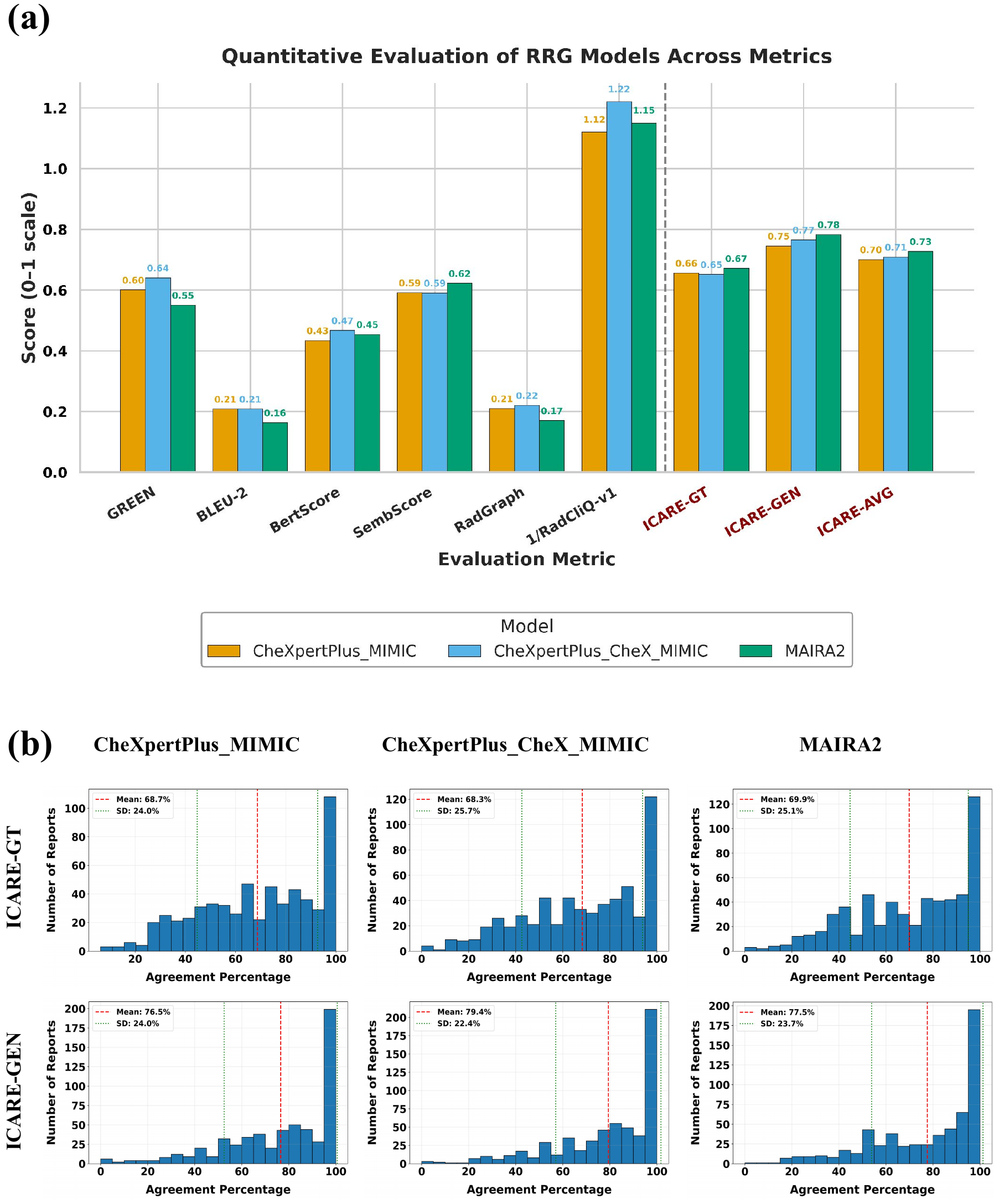}
    \caption{Quantitative results. \textbf{(a)} Comparison of model performance across standard metrics and our agreement-based evaluation (\icare).
    \changes{Our metric captures clinically meaningful differences in model behavior by quantifying both the preservation of reference findings (\icare-GT) and the consistency of additional content (\icare-GEN). MAIRA2 achieves the highest agreement across all variants.} \textbf{(b)} Report-level distribution of \icare-GT and \icare-GEN scores across models and question sources. \icare-GEN scores (agreement on generated-report questions) are generally higher, while \icare-GT scores (agreement on ground-truth questions) show greater variability, reflecting omissions in clinical content in the generated reports.
}
    \label{fig:all_metrics_results}
\end{figure}

\begin{figure}[htbp]
    \centering
    \includegraphics[width=0.9\columnwidth]{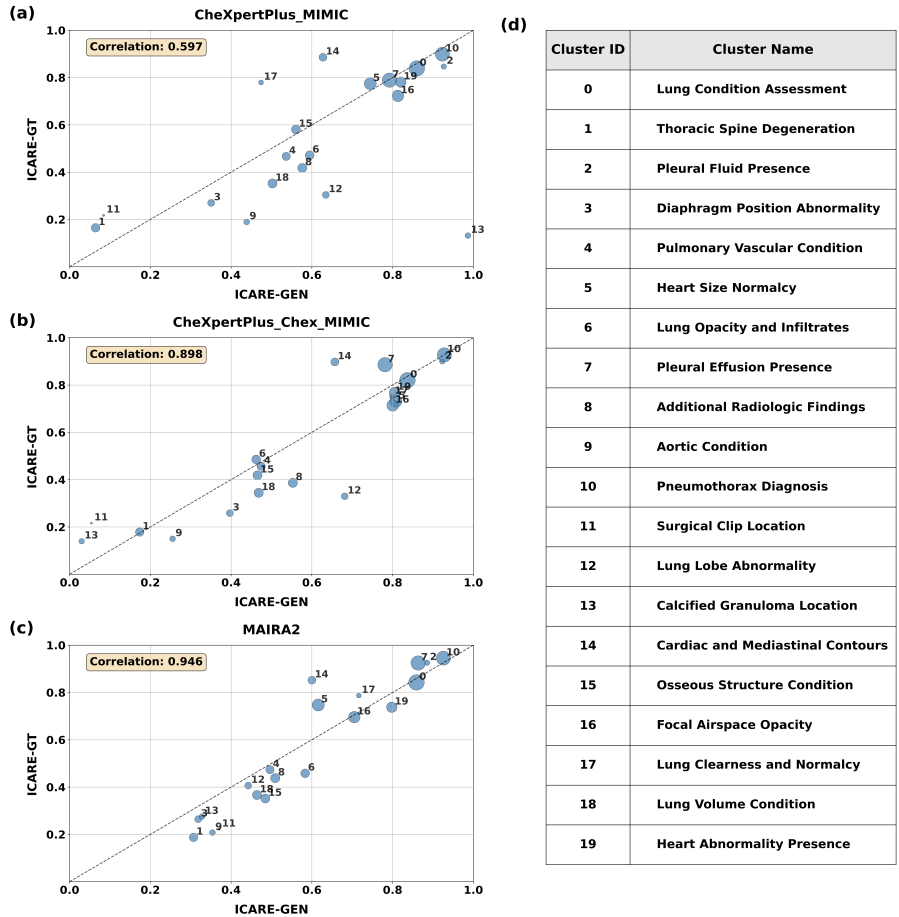} 
\caption{\small \textbf{Cluster-level \icare\ score analysis across RRG models.} 
Panels \textbf{(a)}, \textbf{(b)}, and \textbf{(c)} show scatterplots of \icare\ scores for different clinical clusters across three models. Each point represents a cluster of semantically similar clinical questions. The x-axis shows the \icare-GEN score, i.e., agreement on questions produced from the generated report (reflecting the consistency of added content), while the y-axis shows the \icare-GT score, i.e., agreement on questions produced from the ground-truth report (reflecting preservation of key findings). Point size corresponds to the number of questions in each cluster. Panel \textbf{(d)} provides descriptive cluster names. Clusters below the diagonal, where agreement on ground-truth questions exceeds that on generated-report questions, indicate omission-dominated errors. Clusters above the diagonal reflect hallucination-dominated errors, where unsupported content is introduced. We observe the following: (1) most clusters with more questions, which reflect common clinical concepts, lie near the top right and show high agreement on both axes, indicating strong performance on frequently seen findings; (2) a few clusters with lower agreement on both axes contain fewer questions and typically represent rare or subtle findings in chest X-rays, making them less clinically significant; and (3) the CheXpertPlus variants show several omission-dominated clusters, such as clusters 12 and 13 in CheXpertPlus\_MIMIC and cluster 12 in CheXpertPlus\_CheX\_MIMIC. In contrast, MAIRA2 shows a more compact and balanced distribution of clusters near the diagonal, suggesting greater clinical fidelity. Although all models exhibit some omission and hallucination, MAIRA2 demonstrates the most favorable pattern and appears more reliable overall.}

\label{fig:cluster-analysis-overall}
\end{figure}

\paragraph{Dataset-level results.}

\changes{We evaluated three radiology report generation models using a combination of established automatic metrics and our clinically grounded evaluation framework based on answer agreement. As shown in \cref{fig:all_metrics_results}(a), we compute three \icare\ scores: \icare-GT, using questions produced from the ground-truth report; \icare-GEN, using questions from the generated report; and their average, \icare-AVG. MAIRA2 achieves the highest values across all three, indicating stronger alignment with clinically meaningful content.}

\changes{Beyond separating models, our metric reveals clinically interpretable patterns in model behavior. Across all models, \icare-GT scores (agreement on questions derived from the ground-truth report) are consistently lower than \icare-GEN scores (agreement on questions derived from the generated report). 
This indicates that models are more likely to omit relevant clinical findings than to introduce unsupported content. Traditional metrics generally fail to capture this distinction, whereas our approach enables targeted assessment of both types of errors.}

\changes{While recent report generation models demonstrate promising performance, our findings reveal that they still fall short of reliably capturing the full clinical content of radiologists-written reports. 
These limitations are often obscured by existing evaluation metrics, which tend to emphasize surface-level similarity, predefined entity structures, or learned preferences rather than clinical completeness. 
In contrast, our evaluation provides a structured and interpretable signal that directly reflects clinical fidelity. By linking model performance to specific preserved and omitted findings, our framework offers a complementary and necessary perspective for assessing radiology report generation systems and supports the development of more clinically robust models.}

\paragraph{Report-level results.}
We analyze the distribution of report-level \icare\ scores (\icare-GT and \icare-GEN) across different model variants, as shown in~\cref{fig:all_metrics_results}(b). For each model, we separately evaluate questions generated from ground-truth reports and from generated reports. Our findings reveal that for all models, a large majority of reports achieve high \icare\ scores, indicating substantial clinical similarity between ground-truth and generated reports.

When using ground-truth reports as reference, the distribution of \icare-GT scores is slightly wider, with a small proportion of reports exhibiting lower agreement. In contrast, when using generated reports as reference, \icare-GEN scores are more concentrated toward higher values, suggesting that generated reports may not fully capture the richer clinical details present in ground-truth reports. This pattern is consistent across all evaluated models.

Among the model variants, MAIRA-2 achieves the highest mean report-level \icare-AVG, followed by CheXpertPlus\_CheX\_MIMIC and CheXpertPlus\_MIMIC. These trends are aligned with the dataset-level results and support the observation that ground-truth reports contain richer clinical content compared to automatically generated reports. Overall, the distribution plots confirm that our MCQA-based evaluation captures clinically meaningful differences at the individual report level.

\paragraph{Robustness of our metric.} To evaluate the robustness of our metric, we performed controlled perturbation experiments and stability analyses. Perturbation tests involving random deletion of words demonstrate that agreement scores (\icare\ scores) systematically degrade with increasing report distortion, confirming sensitivity to clinically meaningful content changes (see Extended Dataset~\cref{fig:perturbation-results}). Additionally, \icare\ scores remain highly stable across evaluation seeds, with standard deviations below 1\% at the dataset level 
and low variability at the report level (see Extended Dataset~\cref{fig:report-level-distribution-heatmaps}). At the report level, stability was evaluated by measuring the standard deviation of report-level agreement scores across seeds. Over $85\%$ of reports exhibit a standard deviation of at most $15$, and less than $6\%$ of reports show variability greater than $20$. Reports with very high ($>80\%$) or very low ($<20\%$) mean agreement are especially stable, with more than 95\% maintaining low variability. In contrast, reports with intermediate agreement ($20–80\%$) show slightly higher variability, reflecting lower stability when reports are neither clearly similar nor clearly different. These results highlight that the metric maintains stable performance across different question and answer generation seeds, supporting its robustness for clinical-scale evaluations.

\subsection{Question Categorization and Cluster-level Analysis}

To better understand how clinical content influences evaluation outcomes, we performed a cluster-level analysis of \icare-GT and \icare-GEN scores across different categories of clinical questions. We first collected all unique questions generated across all models, and seeds. Using MedCPT\citep{Jin2023BioCPTCP}, a medical-domain language model, we computed embeddings for each question and applied K-means clustering to group them into 20 semantically coherent clusters. From each cluster, five representative questions were sampled, and LLAMA 3.1 70B language model was prompted to generate a descriptive name for each cluster. These cluster names are shown in Figure~\ref{fig:cluster-analysis-overall}d and reflect a diverse range of clinical concepts. 
Extended Dataset~\cref{tab:mcqa_cluster_examples} presents one sample question from each of the 20 semantic clusters to illustrate the types of clinical questions generated by our MCQ Dataset Generation module.

As shown in Figure~\ref{fig:cluster-analysis-overall}, each cluster is visualized as a point in a scatterplot, where the x-axis shows \icare-GEN scores (agreement on questions produced from the generated report) and the y-axis shows \icare-GT scores (agreement on questions from the ground-truth report). Point size reflects the number of questions in the cluster. Extended Dataset~\cref{tab:ques_categorization_counts} lists the exact number of questions originating from each clinical cluster for the ground-truth and generated reports across all three models, providing a reference for how frequently different clinical concepts were evaluated. We observe substantial variation in \icare\ scores across clinical categories. Clusters corresponding to common findings such as pleural effusion and cardiomegaly, represented by larger points, consistently show high \icare\ scores on both axes, indicating reliable model performance on frequently observed clinical content. In contrast, smaller clusters representing rarer or subtler findings often exhibit lower \icare\ scores on both axes, and their limited size suggests reduced clinical impact.

We also identify systematic error patterns based on the relative position of clusters with respect to the diagonal. Clusters below the diagonal, where \icare-GT exceeds \icare-GEN, indicate omission dominated errors, suggesting that important clinical content present in the ground truth report is often omitted in the generated report. Clusters above the diagonal reflect hallucination dominated errors, where unsupported information appears in the generated report. These omission patterns are more prominent in the CheXpertPlus\_MIMIC and CheXpertPlus\_CheX\_MIMIC models, especially in clusters 12 and 13. In contrast, MAIRA2 exhibits a more compact and balanced cluster distribution near the diagonal, suggesting greater clinical fidelity, fewer severe omissions, and more reliable content generation.

These findings highlight that radiology report generation model performance varies across clinical concepts. Cluster-level evaluation provides a clinically meaningful and interpretable breakdown of model behavior. It reveals which types of content are well preserved, omitted, or hallucinated, offering insight into both strengths and typical failure modes.

\section{Discussion}


We present a clinically grounded and interpretable evaluation framework for assessing radiology report generation (RRG) systems. Our approach, \icare, introduces a dual-agent setup that leverages multiple-choice question answering (MCQA) to compare the clinical content of ground-truth and generated reports. Unlike prior metrics, which are either surface-level (such as BLEU and ROUGE), opaque (such as BERTScore or GREEN), or tied to limited clinical ontologies (such as CheXpert or RadGraph), our method provides a transparent and scalable mechanism for evaluating clinical accuracy and completeness in generated reports.

Using this framework, we evaluated multiple RRG systems, including MAIRA-2 and CheXpertPlus variants. While prior metrics suggested that these models were high-performing, our method reveals that none of them consistently preserve clinically critical information. Even the strongest models frequently miss subtle but important findings. This discrepancy highlights the limitations of existing metrics and shows that they can mask significant clinical deficiencies.
Importantly, our approach does not merely expose that models underperform. It shows \textbf{why} and \textbf{where} they underperform by linking performance to specific question–answer pairs. This allows model developers to pinpoint which aspects of the report (such as certain anatomical regions or conditions) are not being reliably captured, providing a clear target for improvement.

A key advantage of our method is that it produces two agreement scores, \icare-GT and \icare-GEN, that serve as interpretable proxies for precision and recall. Agreement on questions produced from the ground-truth report (\icare-GT) reflects how well the generated report preserves clinically relevant information, acting as a proxy for precision. Agreement on questions derived from the generated report (\icare-GEN) reflects whether the content introduced by the model is supported by the reference report, serving as a proxy for recall. This structure allows us to distinguish between different types of errors by identifying whether a model tends to omit important findings or introduce unsupported information. We observe a consistent pattern across models: agreement on generated-report questions (\icare-GEN) is generally higher than on ground-truth questions (\icare-GT), suggesting that omissions are more common than hallucinations. This pattern is especially evident in the CheXpertPlus variants, while MAIRA-2 demonstrates more balanced behavior. These findings underscore the importance of clinically informed evaluation methods that provide deeper insight into the quality of generated content beyond what is captured by traditional metrics.

Beyond this decomposition, our framework offers interpretability by organizing questions into clinical categories. Using semantic clustering, we show that models perform unevenly across clinical concepts. Common findings such as pleural effusion and cardiomegaly are often retained, while more complex or infrequent findings such as thoracic spine changes, or subtle lung volume abnormalities are frequently omitted. This cluster-level analysis moves evaluation beyond a single score and provides actionable insight into model behavior, enabling targeted improvement.

To assess the clinical validity of our framework, we conducted a human evaluation in which radiologists were asked to compare pairs of model-generated reports. When the \icare\ score difference between reports was large, clinicians consistently preferred the report with the higher score, suggesting that our metric captures distinctions that matter in clinical decision-making. When the score difference was small, clinicians frequently selected “Can’t say.” These cases often involved reports that were either similarly sufficient or similarly lacking in relevant clinical content. Our metric reflects this same pattern. When two reports are close in score, it typically indicates that they convey similar levels of information, whether strong or weak. This correspondence between clinician behavior and \icare\ score gaps supports the conclusion that our method not only correlates with expert judgment but also meaningfully captures uncertainty in cases where report quality is difficult to distinguish.

These findings carry several implications. First, our framework provides a foundation for future model development. It not only evaluates performance but also guides improvement by exposing specific areas of weakness. Second, RRG systems remain far from the clinical reliability required for standalone use.  Third, while our evaluation focuses on chest X-ray reports, the framework is not limited to this domain. It can be extended to other imaging modalities such as CT and MRI, as well as to other clinical report or note generation tasks beyond radiology, such as pathology or discharge summaries. Evaluating its applicability and effectiveness in these broader contexts is an important direction for future work. Finally, the framework enables post-deployment monitoring even in the absence of ground-truth reports, by tracking the answerability of curated clinical questions over time.

In summary, our evaluation framework offers a clinically meaningful, interpretable, and scalable alternative to existing metrics. By revealing not only how well a model performs but also why it fails, we provide a framework that supports safer model development, more transparent evaluation, and more trustworthy clinical AI systems.

\section{Method}

\subsection{Overview}

\noindent
We propose an evaluation framework, \icare, consisting of two parallel report-aware agents designed to assess the clinical similarity between a generated radiology report and its corresponding ground-truth report. Agent\textsubscript{GT} receives the ground-truth report as input, while Agent\textsubscript{GEN} receives the generated report. Within each agent, two steps are performed independently: first, the agent generates multiple-choice questions based on its assigned report; second, it answers both its own and the other agent’s questions using only its input report. After both agents have answered the questions, we perform an external answer agreement evaluation that separately measures agreement on questions originating from the ground-truth report and those originating from the generated report. 
Agreement on answers to ground-truth questions assesses whether the generated report preserves clinically important information, and is denoted as \icare-GT, serving as a proxy for precision. Agreement on answers to generated-report questions assesses whether any additional information introduced is clinically consistent with the ground-truth report, and is denoted as \icare-GEN, serving as a proxy for recall. The average of these two scores, referred to as \icare-AVG, summarizes overall alignment.
This structure proceeds in three stages: MCQ Dataset Generation within each agent, MCQ Answer Generation within each agent, and Answer Agreement Evaluation across agents, as illustrated in Figure~\ref{fig:panel1}.

\subsection{Step 1: MCQ Dataset Generation (within each agent)}

To evaluate the clinical content captured in each report, we first generate multiple-choice questions (MCQs) independently for the ground-truth and generated reports. Each report is processed by a dedicated agent:

\begin{itemize}
    \item Agent\textsubscript{GT} receives the ground-truth report \( R_{\text{GT}} \),
    \item Agent\textsubscript{GEN} receives the generated report \( R_{\text{GEN}} \).
\end{itemize}

Using a large language model, each agent generates a set of \( n \) MCQs that are specific to the clinical details described in its assigned report. The generated questions are designed to probe various clinical aspects, such as the location, severity, or presence of findings. We use the LLAMA 3.1 70B~\citep{Dubey2024TheL3} language model for question generation, guided by the prompt format illustrated in Extended Dataset~\cref{fig:prompts}.

Each MCQ consists of:
\begin{itemize}
\vspace{-5pt}
    \item A question prompt \( Q \) related to the report content,
    \item Four answer choices \( \{a^1, a^2, a^3, a^4\} \),
    \item A correct answer \( a^* \in \{a^1, a^2, a^3, a^4\} \).
\end{itemize}

Formally, the collections of questions generated by each agent are denoted as:

\begin{align*}
    Q_{\text{GT}} &= \{Q_{\text{GT},i}\}_{i=1}^n, \\
    Q_{\text{GEN}} &= \{Q_{\text{GEN},j}\}_{j=1}^n.
\end{align*}

Questions are generated from both \(R_{\text{GT}} \) and \( R_{\text{GEN}} \) to ensure that the evaluation captures the unique information present in both reports. This dual-reference approach ensures that the evaluation framework identifies whether \( R_{\text{GEN}} \) provides new, relevant information or introduces hallucinated details.

\paragraph{Filtering clinically meaningful questions.}

\begin{table}[t]
\centering
\adjustbox{max width=\columnwidth}{%
\begin{tabular}{@{} l | l | l | c @{}}
\toprule
\makecell{\textbf{Ans w/ report}} & \makecell{\textbf{Ans w/o report}} & \makecell{\textbf{Correctness}} & \makecell{\textbf{Selection criteria}} \\ \midrule
correct (a) & correct (a) & Both correct and in agreement & \xmark \\
incorrect (a) & incorrect (a) & Both incorrect and in agreement & \xmark \\ 
correct (a) & incorrect (b) & w/ report correct. No agreement & \cmark \\ 
incorrect (a) & correct (a) & w/o report correct. No agreement & \xmark \\
\bottomrule
\end{tabular}
}
\caption{Selection criteria derived from all possible options to identify questions that require reports for generating answers.}
\label{tab:filtering_criteria}
\end{table}



To ensure that the generated questions require access to the report content rather than general medical knowledge, a filtering step is applied. Using the same LLAMA 3.1 70B model\cite{Dubey2024TheL3}, we answer each question both with and without access to the report. We compute the following:

$ P_{\text{with}}(Q_k, R): $ 
accuracy in predicting the correct answer choice $a_k^*$ for question $Q_k$ using report $R$ and 

$ P_{\text{without}}(Q_k): $ accuracy in predicting the correct answer choice $a_k^*$ for question $Q_k$ without using report $R$, using general knowledge of the LLAMA model.

A question is retained if and only if

\[
P_{\text{with}}(Q_k, R) = 1, \quad P_{\text{without}}(Q_k) = 0,
\]

\noindent meaning that the question can be answered correctly only when the report is available. Table~\ref{tab:filtering_criteria} summarizes the possible answer scenarios and the corresponding selection criteria. This ensures that only report-dependent questions are retained.
Formally, the filtered sets of questions \( Q_{\text{filtered,GT}} \) and \( Q_{\text{filtered,GEN}} \) are defined as:
\noindent\scalebox{0.93}{%
\begin{minipage}{\linewidth}
\begin{align*}
    Q_{\text{filtered,GT}} &= 
    \big\{ Q_{\text{GT},i} \mid P_{\text{with}}(Q_{\text{GT},i}, R_{\text{GT}}) = 1, \\
    & \qquad \text{and } P_{\text{without}}(Q_{\text{GT},i}) = 0 \big\}_{i=1}^n, \\
    Q_{\text{filtered,GEN}} &= 
    \big\{ Q_{\text{GEN},j} \mid P_{\text{with}}(Q_{\text{GEN},j}, R_{\text{GEN}}) = 1, \\
    & \qquad \text{and } P_{\text{without}}(Q_{\text{GEN},j}) = 0 \big\}_{j=1}^n.
\end{align*}
\end{minipage}%
}

The resulting filtered sets, \( Q_{\text{filtered,GT}} \) and \( Q_{\text{filtered,GEN}} \), contain clinically meaningful, report-dependent questions, and form the MCQ datasets used for subsequent answer generation and evaluation.


As summarized in Extended Dataset Table~\ref{tab:filtering_results}, we find that while accuracy is high when questions are answered using the report ($\sim$98\%), a substantial proportion of questions ($\sim$22\% to 33\%) can also be answered correctly without access to the report. This suggests that many generated questions reflect general medical knowledge rather than report-specific content. To isolate clinically meaningful, report-dependent questions, we apply a strict filtering criterion that retains only those answered correctly with the report but incorrectly without it. After filtering, approximately 25\% to 31\% of questions are retained across the three models. On average, this yields 8 to 13 report-dependent questions per report for both ground-truth and generated reports. The resulting filtered datasets, $Q_{\text{filtered,GT}}$ and $Q_{\text{filtered,GEN}}$, include diverse clinical queries related to the location, severity, and presence of findings. Examples of such questions are shown in ~\cref{fig:panel1}b.

\paragraph{Bias mitigation.}
Following filtering, we observed that the correct answer choices in the MCQ datasets exhibited biases toward specific options (e.g., a predominance of choice "B"). To mitigate this, we randomly shuffle the order of answer choices \( \{a_k^1, a_k^2, a_k^3, a_k^4\} \) both before and after filtering. This shuffling yields a more balanced distribution across answer choices, ensuring that the evaluation is not influenced by systematic biases in answer positioning.

\subsection{Step 2: MCQ Answer Generation (within each agent)}

Following MCQ dataset generation and filtering, each agent independently answers both sets of filtered questions using the LLAMA 3.1 70B\citep{Dubey2024TheL3}:

\begin{itemize}
    \item Agent\textsubscript{GT} uses \( R_{\text{GT}} \) to answer questions from both \( Q_{\text{filtered,GT}} \) and \( Q_{\text{filtered,GEN}} \),
    \item Agent\textsubscript{GEN} uses \( R_{\text{GEN}} \) to answer questions from both \( Q_{\text{filtered,GT}} \) and \( Q_{\text{filtered,GEN}} \).
\end{itemize}

Formally, this process results in four sets of answers:

\begin{itemize}
    \item \( A_{\text{GT}}(Q_{\text{filtered,GT}}) \): answers by Agent\textsubscript{GT} to questions from \( Q_{\text{filtered,GT}} \),
    \item \( A_{\text{GT}}(Q_{\text{filtered,GEN}}) \): answers by Agent\textsubscript{GT} to questions from \( Q_{\text{filtered,GEN}} \),
    \item \( A_{\text{GEN}}(Q_{\text{filtered,GT}}) \): answers by Agent\textsubscript{GEN} to questions from \( Q_{\text{filtered,GT}} \),
    \item \( A_{\text{GEN}}(Q_{\text{filtered,GEN}}) \): answers by Agent\textsubscript{GEN} to questions from \( Q_{\text{filtered,GEN}} \).
\end{itemize}

Each answer is generated solely based on the agent's assigned report, ensuring that the agents operate independently and that all answers reflect only the information contained within the respective reports.

\subsection{Step 3: Answer Agreement Evaluation (across agents)}

Once both agents have produced their answers, we perform an external answer agreement evaluation to measure the clinical similarity between the ground-truth and generated reports. Agreement is computed separately for questions originating from the ground-truth report, denoted by \icare-GT, and those originating from the generated report, denoted by \icare-GEN.

\paragraph{Report-level agreement.}
We compute agreement scores separately for each report, providing a fine-grained, interpretable view of performance.  
For a given report \( r \), the report-level agreement scores are defined as:

\[
S_{\text{GT},r} = \frac{1}{|Q_{\text{filtered,GT},r}|} \sum_{Q_k \in Q_{\text{filtered,GT},r}} \mathbb{I}(A_{\text{GT}}(Q_k) = A_{\text{GEN}}(Q_k)),
\]
\[
S_{\text{GEN},r} = \frac{1}{|Q_{\text{filtered,GEN},r}|} \sum_{Q_k \in Q_{\text{filtered,GEN},r}} \mathbb{I}(A_{\text{GT}}(Q_k) = A_{\text{GEN}}(Q_k)),
\]
where \( Q_{\text{filtered,GT},r} \) and \( Q_{\text{filtered,GEN},r} \) denote the subsets of questions associated with report \( r \). These report-level scores provides a similarity score based on the number of questions that are answered in agreement for every report.

\paragraph{Dataset-level agreement.}
At the dataset level, we aggregate agreement across all reports by comparing the answers produced by the two agents for each question. For questions originating from the ground-truth report (\( Q_{\text{filtered,GT}} \)), the dataset-level agreement score \( S_{\text{GT}} \) is computed as:

\[
S_{\text{GT}} = \frac{1}{|Q_{\text{filtered,GT}}|} \sum_{Q_k \in Q_{\text{filtered,GT}}} \mathbb{I}(A_{\text{GT}}(Q_k) = A_{\text{GEN}}(Q_k)),
\]

where \( \mathbb{I}(\cdot) \) is the indicator function that equals 1 if the two agents provide the same answer and 0 otherwise.

Similarly, for questions originating from the generated report (\( Q_{\text{filtered,GEN}} \)), the dataset-level agreement score \( S_{\text{GEN}} \) is computed as:

\[
S_{\text{GEN}} = \frac{1}{|Q_{\text{filtered,GEN}}|} \sum_{Q_k \in Q_{\text{filtered,GEN}}} \mathbb{I}(A_{\text{GT}}(Q_k) = A_{\text{GEN}}(Q_k)).
\]
These scores provide a detailed quantitative measure of the similarity between \( R_{\text{GT}} \) and \( R_{\text{GEN}} \). The per-report \icare\ scores, \( S_{\text{GT},r} \) and \( S_{\text{GEN},r} \), allow us to analyze similarity on a report-by-report basis, while the dataset-level \icare\ scores, \( S_{\text{GT}} \) and \( S_{\text{GEN}} \), capture aggregate agreement across all questions.

\paragraph{Interpretation of agreement scores.}
\begin{itemize}
    \item Agreement on questions generated from the ground-truth report (\icare-GT) reflects the degree to which clinically important information is preserved in the generated report (analogous to precision).
    \item Agreement on questions generated from the generated report (\icare-GEN) reflects the degree to which additional information introduced in the generated report is clinically consistent with the ground-truth (analogous to recall).
\end{itemize}

This dual evaluation at both dataset and report levels enables both global model assessment and detailed, interpretable analysis of individual reports.







\subsection{Summary}

Our three-stage, agent-based \icare\ framework captures semantic understanding through structured question answering, provides interpretability by linking scores to specific question–answer pairs, and ensures scalability through full automation of question generation, answering, and evaluation. By assessing agreement separately on questions derived from the ground-truth and generated reports, our method offers a clinically grounded, interpretable, and robust framework for evaluating radiology report generation systems.

\backmatter

\begin{appendices}
\renewcommand{\figurename}{Extended Data Fig.}
\renewcommand{\tablename}{Extended Data Table}

\section{Extended Data}
\paragraph{Dataset.} 

We utilize the IU X-ray dataset for our experiments. The IU X-ray dataset contains chest radiographs paired with corresponding radiology reports, covering a variety of thoracic findings. For each study, we consider the associated frontal and lateral chest radiographs along with the accompanying report. This setup enables diverse evaluation settings, allowing assessment of automatically generated reports against clinically written ground-truth reports. For our experiments, we focus on candidate reports generated by multiple model variants, including CheXpertPlus\_MIMIC, CheXpertPlus\_CheXpertPlus\_MIMIC, and MAIRA-2, using both frontal and lateral images from the IU X-ray dataset. Our MCQA framework generates questions independently from ground-truth and generated reports, enabling precise measurement of clinical similarity through answer agreement. This setup allows us to analyze both dataset-level and report-level agreement scores, and to investigate the sensitivity of the evaluation to controlled perturbations applied to the reports.

\paragraph{Experimental setup.} 

\begin{figure*}[t]
\centering
\includegraphics[width=\textwidth]{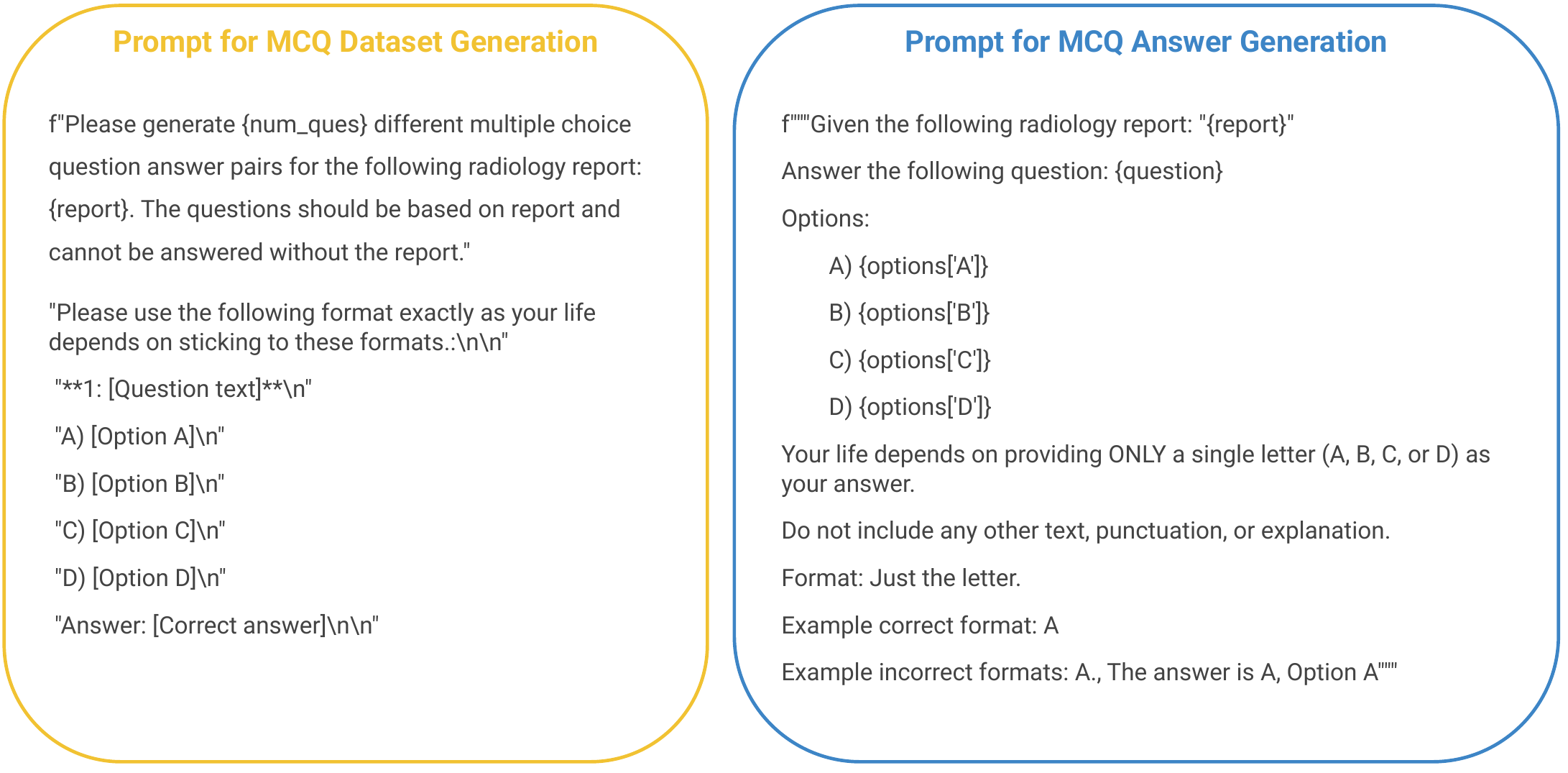}  
\caption{Prompts given to the language model for the MCQ Dataset Generation and MCQ Answer Generation.}  
\label{fig:prompts}  
\end{figure*}

Extended Dataset~\cref{fig:prompts} illustrates the prompts used to instruct the language model during both the dataset generation and MCQA evaluation stages. For the dataset generation, we employed specific prompts to generate multiple-choice questions. During the MCQA evaluation stage, we used a separate prompt to generate answers for computing the similarity between reports.

In the filtering step, we utilized the MCQA evaluation prompt when generating answers using the report. For generating answers without the report, we omitted the line ``Given the following radiology report: \{report\}'' from this prompt. This approach allowed us to differentiate between questions that genuinely required report-specific knowledge and those that could be answered with general medical understanding.

Throughout all stages of the experiment, including dataset generation, filtering, and MCQA evaluation, we consistently used the LLAMA 3.1 language model to ensure uniformity in our approach.

\begin{figure*}[t]
\centering
\includegraphics[width=\textwidth]{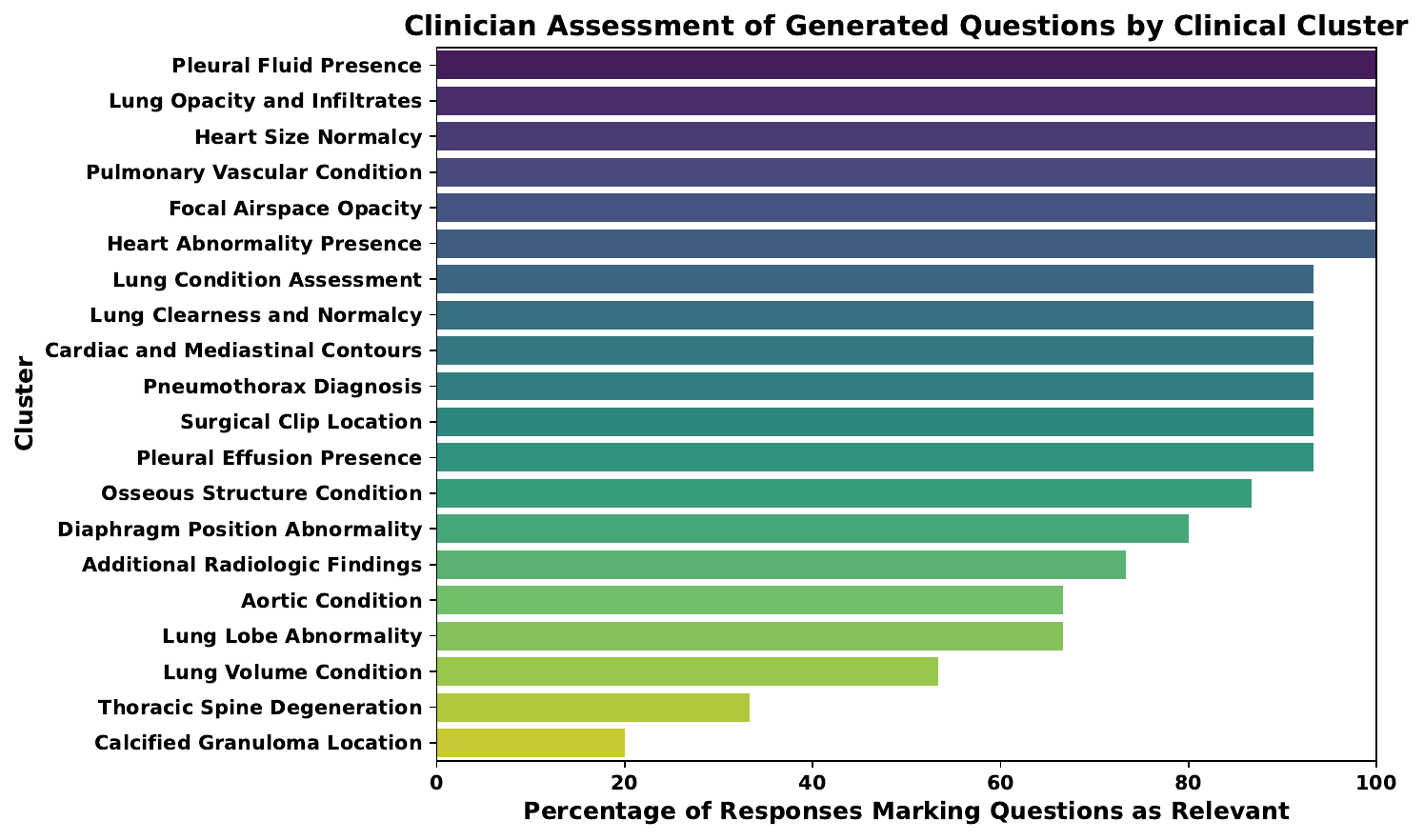}  
\caption{\textbf{Clinical relevance of generated questions across semantic clusters.}
Each bar shows the percentage of clinician responses indicating that questions in a given cluster were appropriate to ask and answerable using only the information typically found in chest X-ray reports. Clusters focused on commonly observed, radiographically accessible findings, such as pleural effusion, lung opacity, and heart size, received the highest endorsement. In contrast, clusters involving subtler or less routinely documented findings, like thoracic spine changes and calcified granulomas, received lower agreement. These results support the clinical validity of our question generation process and highlight which content categories are most compatible with chest X-ray interpretation.
}  
\label{fig:human_eval_ques_with_clusters}  
\end{figure*}

\begin{figure*}[t]
    \centering

    \hspace{0.06\textwidth}
    \begin{minipage}{0.3\textwidth}
        \centering
        {\tiny  \textbf{CheXpertPlus\_MIMIC}}
    \end{minipage}
    \hfill
    \begin{minipage}{0.3\textwidth}
        \centering
        {\tiny \textbf{CheXpertPlus\_CheX\_MIMIC}}
    \end{minipage}
    \hfill
    \begin{minipage}{0.3\textwidth}
        \centering
        {\tiny  \textbf{MAIRA-2}}
    \end{minipage}

    \vspace{0.3cm}

    \begin{minipage}{0.04\textwidth}
        \centering
        \rotatebox{90}{\tiny \textbf{Dataset-level}}
    \end{minipage}%
    \hspace{0.01\textwidth}
    \begin{minipage}{0.93\textwidth}
        \centering
        \begin{subfigure}{0.3\textwidth}
            \includegraphics[width=\linewidth]{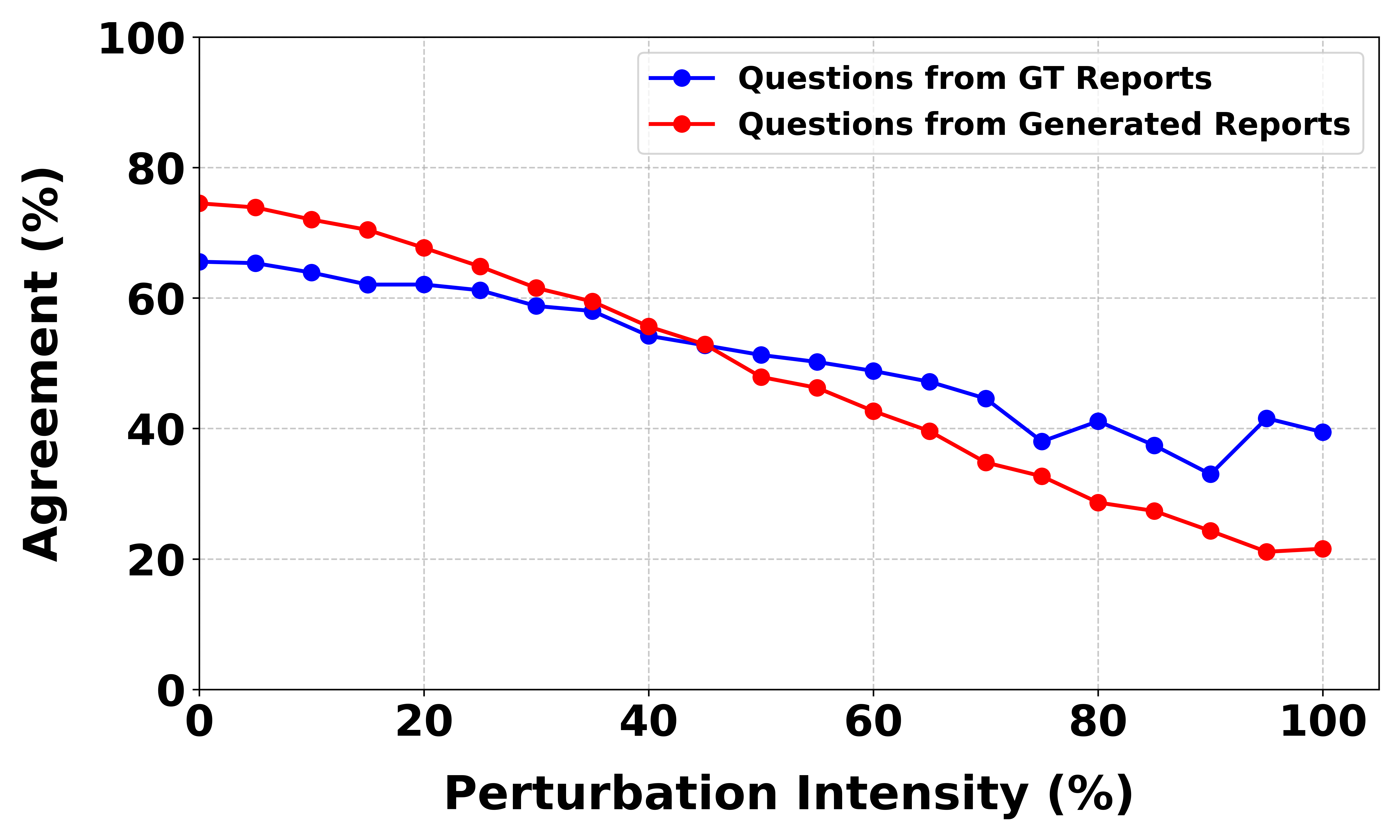}
        \end{subfigure}
        \hfill
        \begin{subfigure}{0.3\textwidth}
            \includegraphics[width=\linewidth]{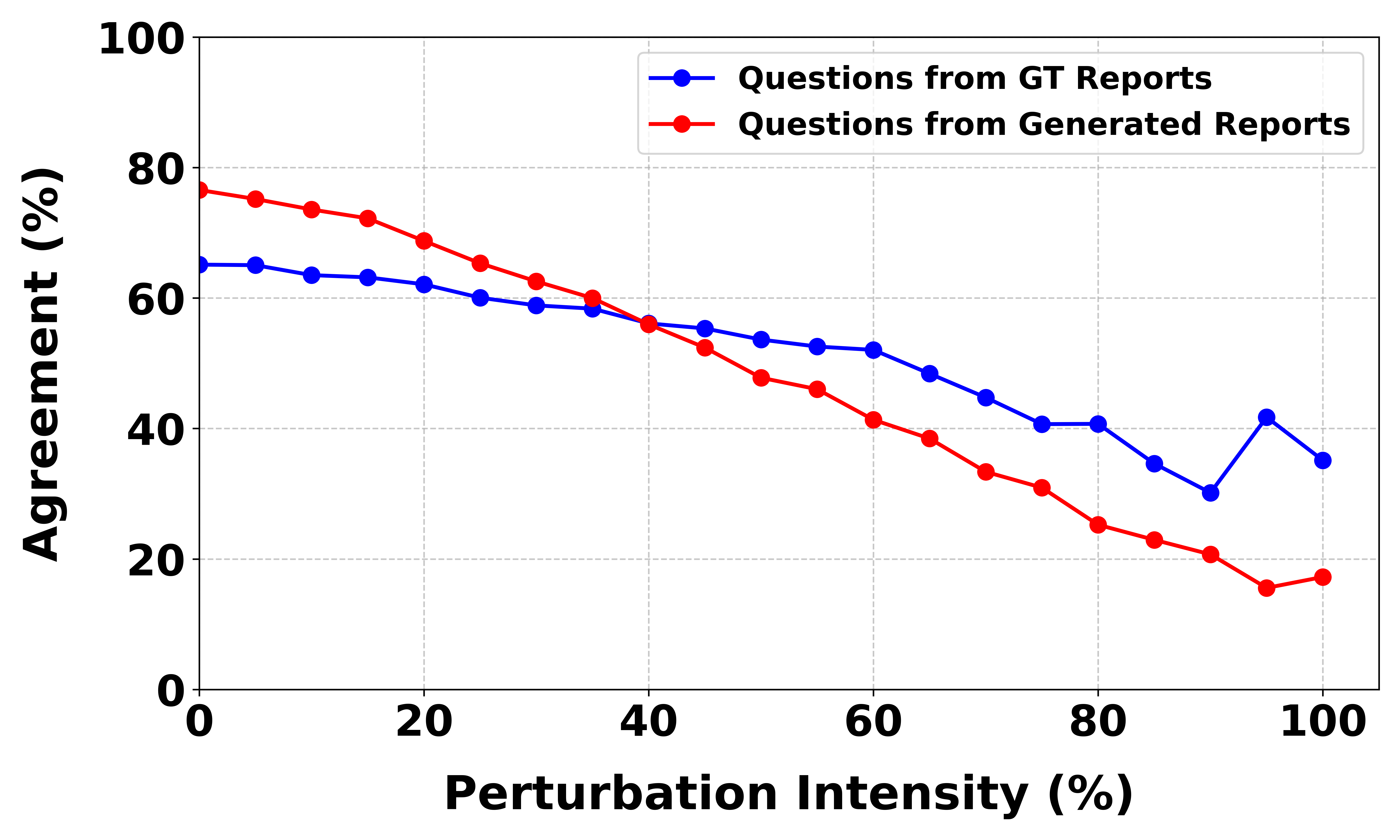}
        \end{subfigure}
        \hfill
        \begin{subfigure}{0.3\textwidth}
            \includegraphics[width=\linewidth]{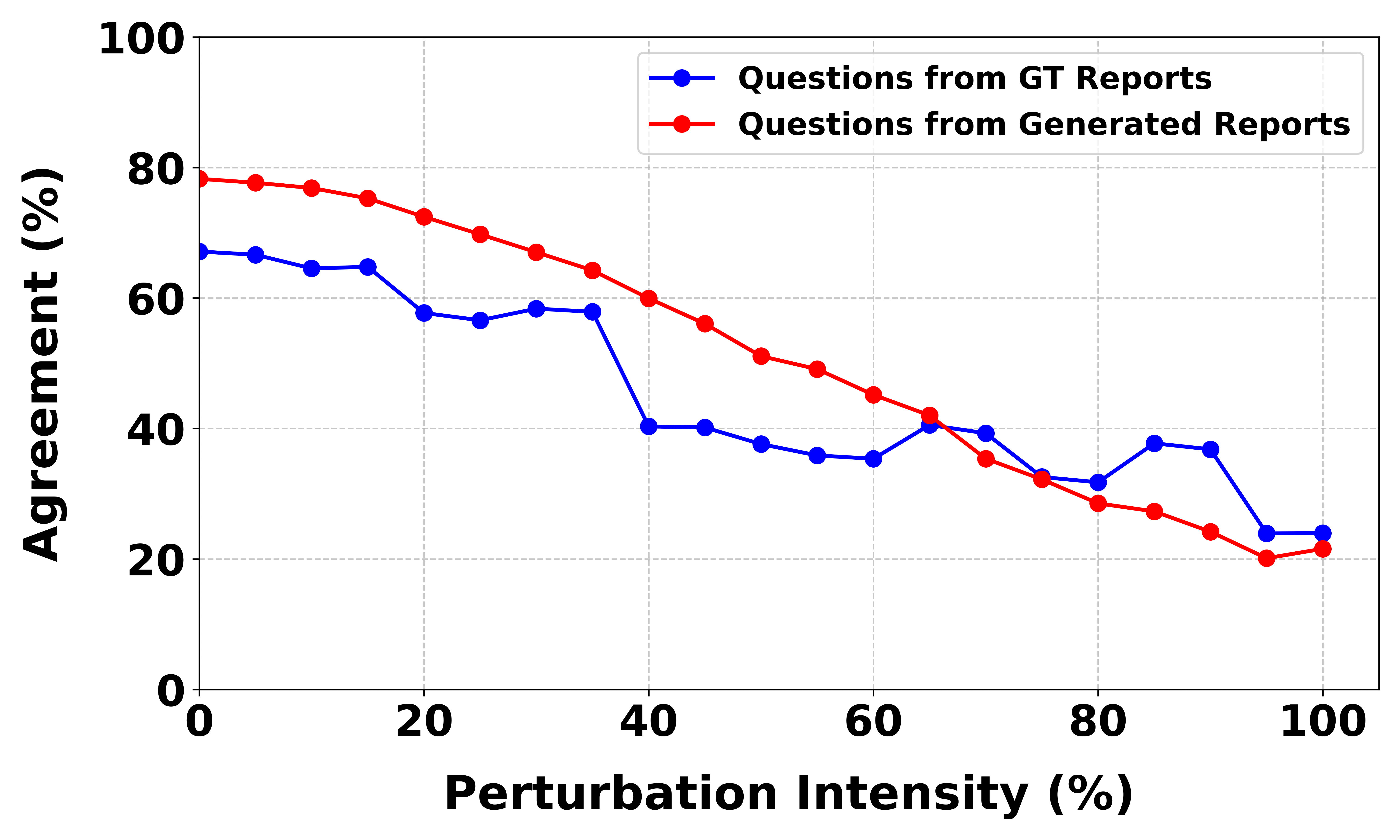}
        \end{subfigure}
    \end{minipage}

    \vspace{0.3cm}

    \begin{minipage}{0.04\textwidth}
        \centering
        \rotatebox{90}{\tiny \textbf{Report-level}}
    \end{minipage}%
    \hspace{0.01\textwidth}
    \begin{minipage}{0.93\textwidth}
        \centering
        \begin{subfigure}{0.3\textwidth}
            \includegraphics[width=\linewidth]{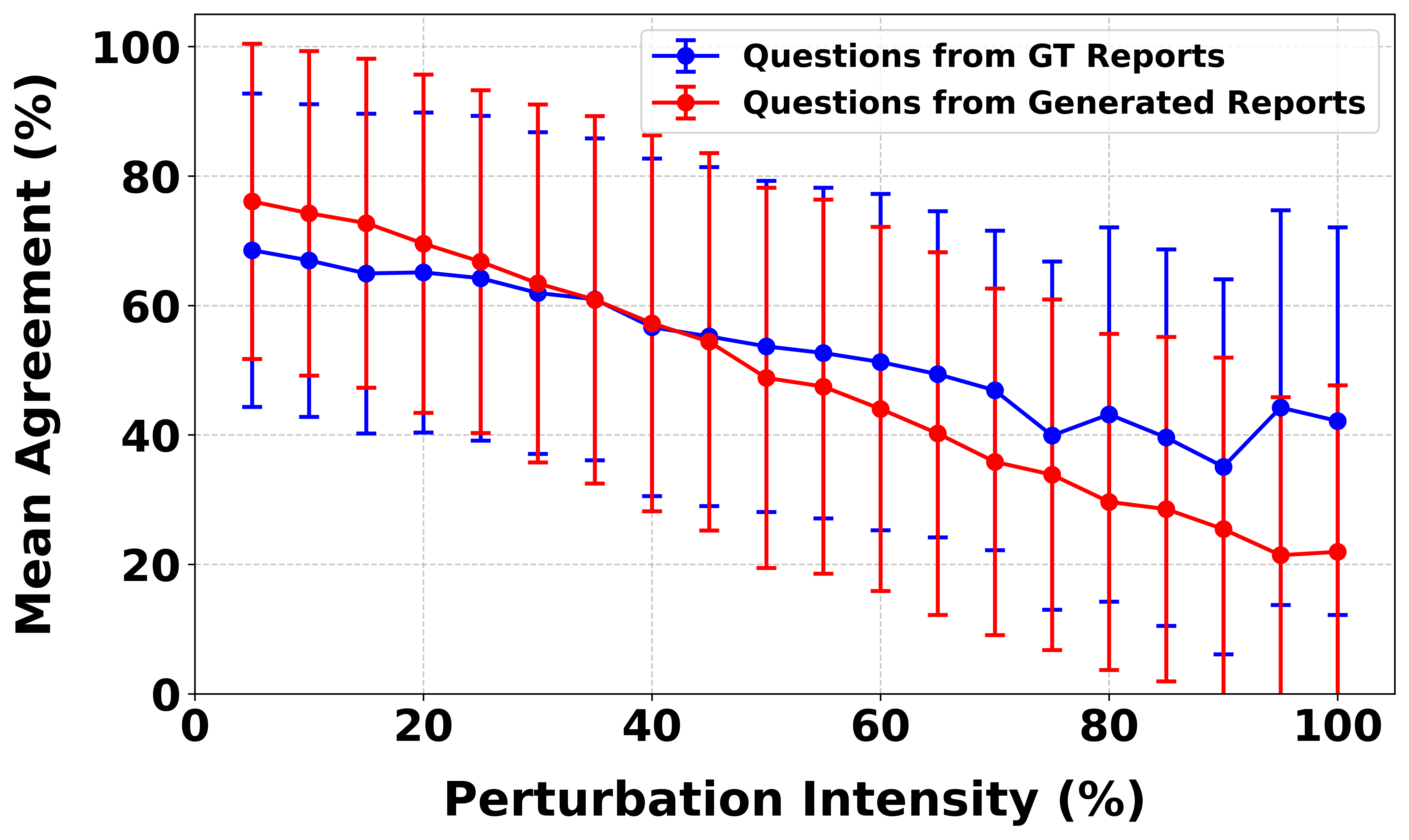}
        \end{subfigure}
        \hfill
        \begin{subfigure}{0.3\textwidth}
            \includegraphics[width=\linewidth]{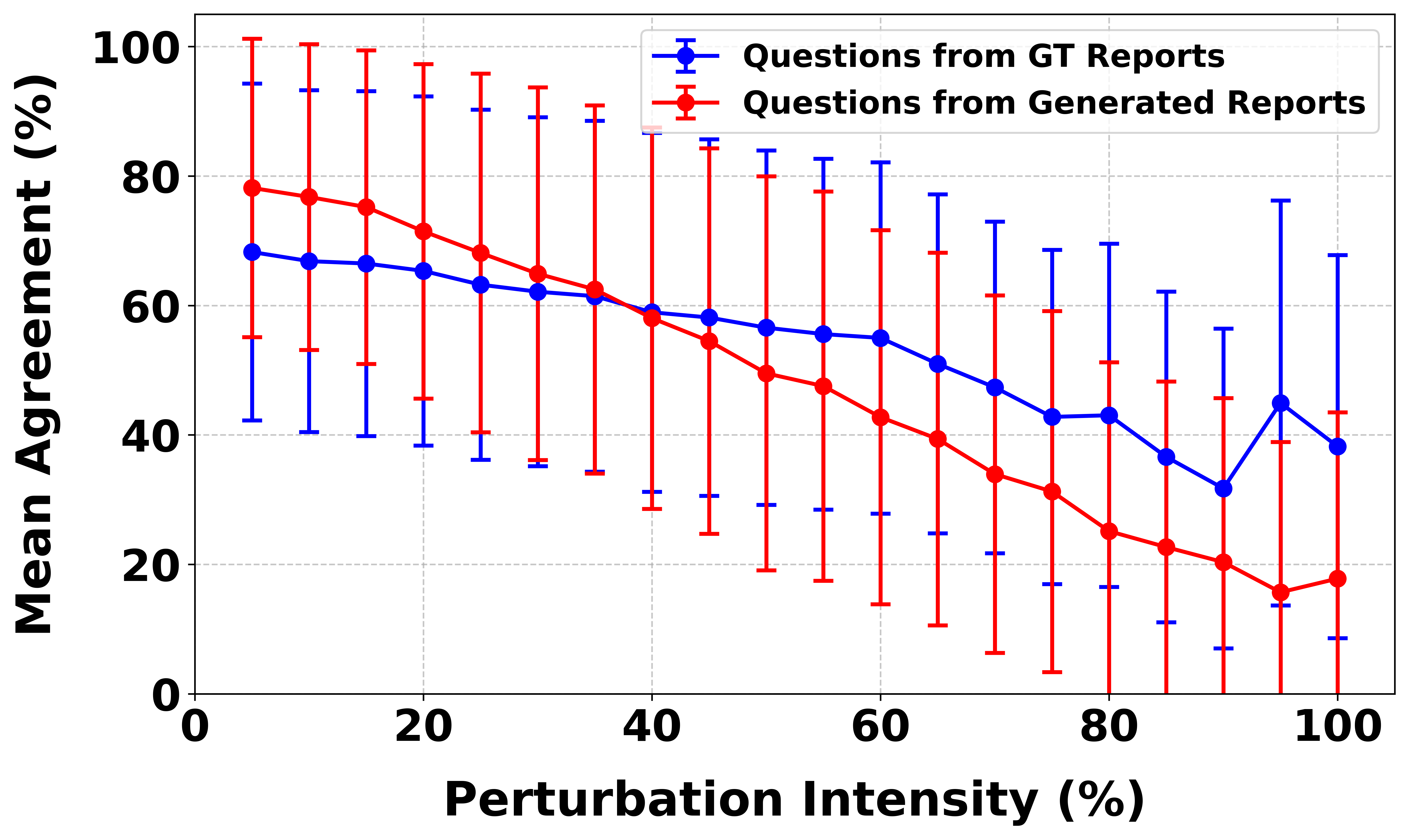}
        \end{subfigure}
        \hfill
        \begin{subfigure}{0.3\textwidth}
            \includegraphics[width=\linewidth]{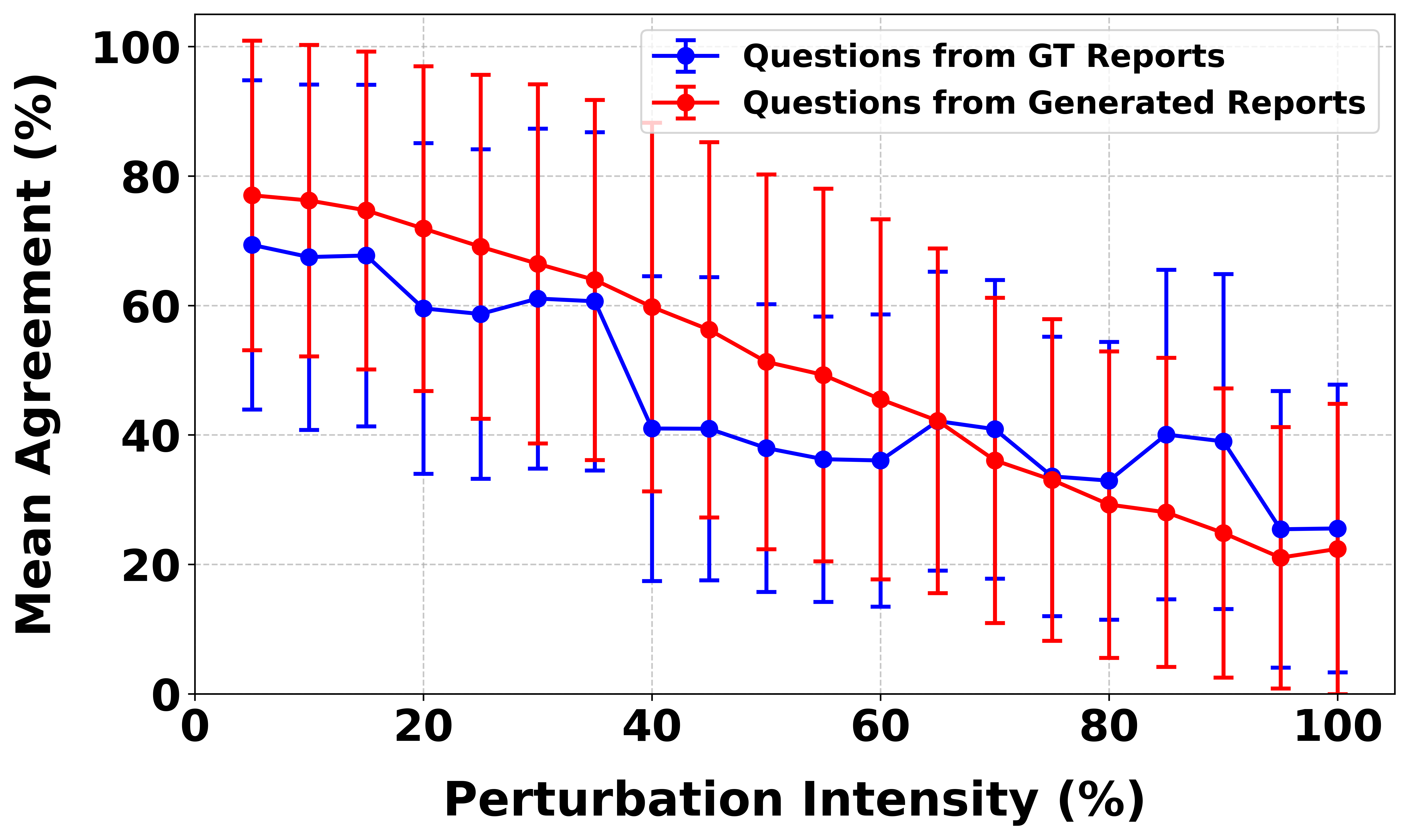}
        \end{subfigure}
    \end{minipage}

    \caption{Perturbation experiments at word level. Top row shows dataset-level agreement percentages(\icare\ scores) versus perturbation intensity. Bottom row shows report-level agreement percentages(\icare\ scores). Each column corresponds to a different model. Blue: questions from ground-truth reports; Red: questions from generated reports. Agreement percentages decrease as perturbation intensity increases.}
    \label{fig:perturbation-results}
\end{figure*}

\begin{figure*}[t]
    \centering

    \begin{minipage}{0.04\textwidth}
        \centering
        \rotatebox{90}{\small \textbf{ICARE-GT}}
    \end{minipage}%
    \hspace{0.01\textwidth}
    \begin{minipage}{0.93\textwidth}
        \centering
        \includegraphics[width=\linewidth]{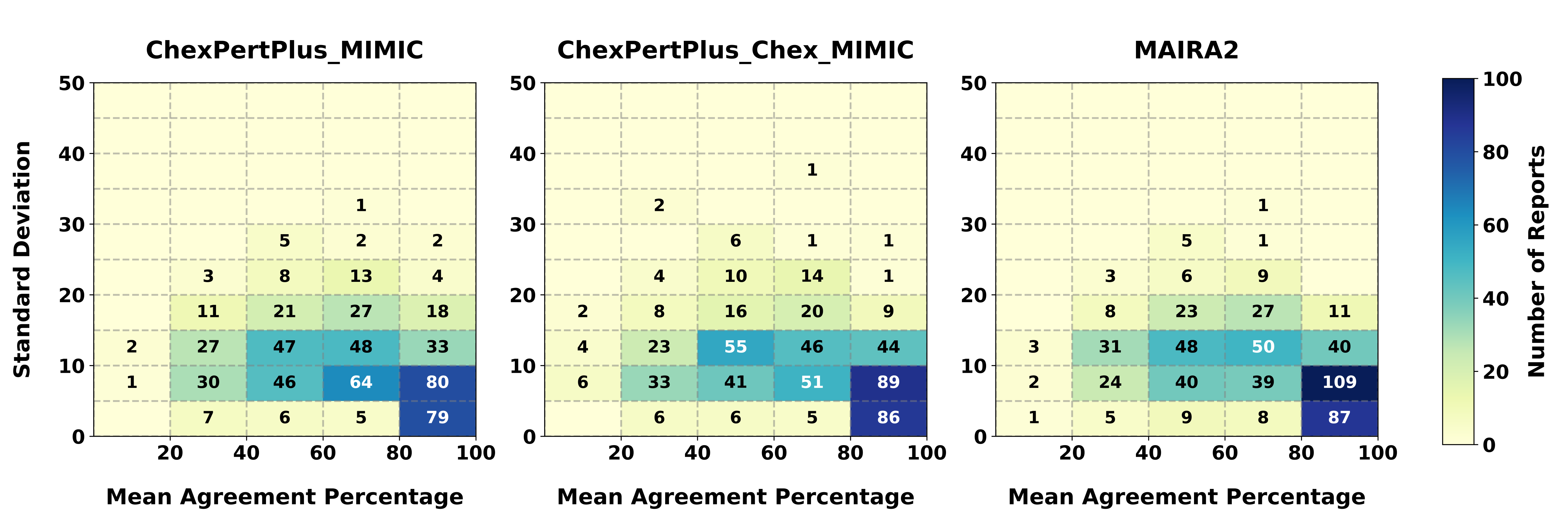}  
    \end{minipage}


    \begin{minipage}{0.04\textwidth}
        \centering
        \rotatebox{90}{\small \textbf{ICARE-GEN}}
    \end{minipage}%
    \hspace{0.01\textwidth}
    \begin{minipage}{0.93\textwidth}
        \centering
        \includegraphics[width=\linewidth]{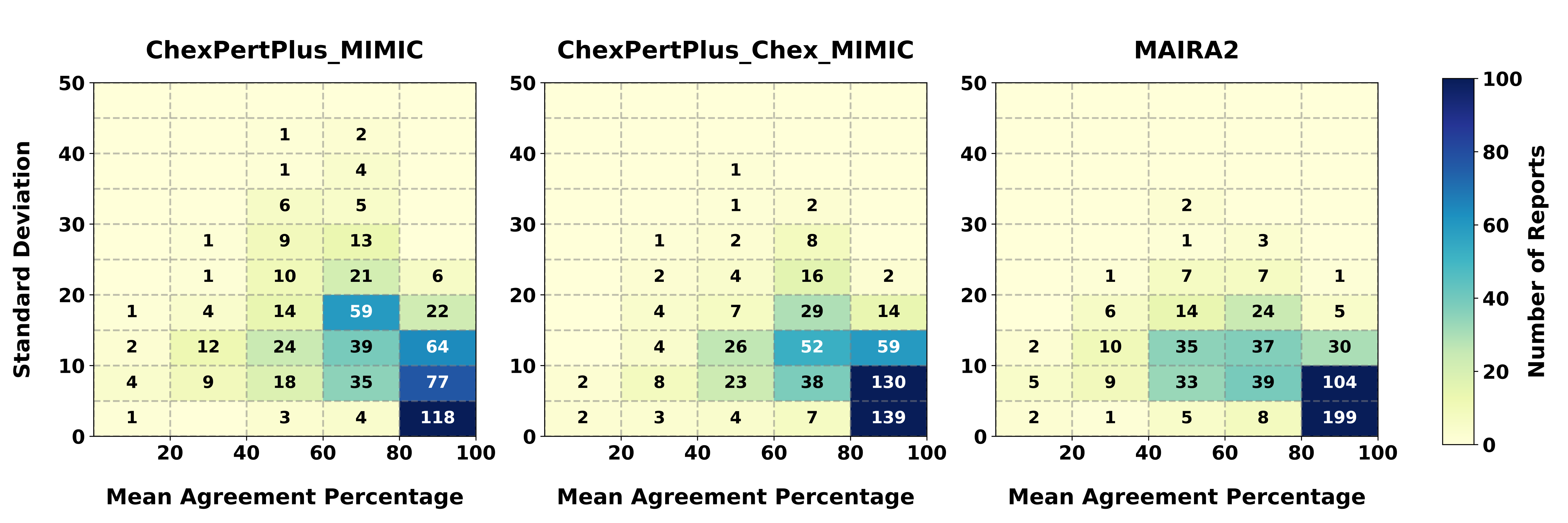}  
    \end{minipage}

    \caption{Distribution of report-level \icare\ scores for different model variants. Top row shows agreement distributions for questions produced from the generated reports(\icare-GEN), while the bottom row shows distributions for questions generated from the ground-truth reports(\icare-GT). Each column corresponds to a different model. The x-axis represents the agreement percentage(\icare\ score), and the y-axis represents the number of reports falling within each agreement range. Overall, the majority of reports achieve high agreement scores, particularly for questions produced from generated reports, while agreement scores for questions based on ground-truth reports exhibit slightly greater variability.}
    \label{fig:report-level-distribution-heatmaps}
\end{figure*}

\begin{table*}[t]
\centering
\adjustbox{max width=\textwidth}{%
\begin{tabular}{@{} l | ll | ll | ll @{}}
\toprule
\textbf{Metric} 
& \multicolumn{2}{c|}{\makecell{\textbf{ChexPertPlus}\\\textbf{MIMIC}}} 
& \multicolumn{2}{c|}{\makecell{\textbf{ChexPertPlus}\\\textbf{Chex\_MIMIC}}}
& \multicolumn{2}{c}{\textbf{MAIRA2}} \\
\cmidrule(l){2-3} \cmidrule(l){4-5} \cmidrule(l){6-7}
& \textbf{Gen.} & \textbf{GT} 
& \textbf{Gen.} & \textbf{GT} 
& \textbf{Gen.} & \textbf{GT} \\ \midrule
\rowcolor{gray!10}
Accuracy with report & 98.29\% & 98.00\% & 98.58\% & 97.95\% & 98.17\% & 97.94\% \\
Accuracy without report & 76.61\% & 69.07\% & 73.03\% & 68.97\% & 65.93\% & 68.66\% \\
\rowcolor{gray!10}
Ques (\#) correct w/ report, incorrect w/o 
& 5271/23600 & 7008/23600 & 6149/23600 & 7022/23600 & 7738/23600 & 7089/23600 \\
Ques (\%) correct w/ report, incorrect w/o
& 22.33\% & 29.69\% & 26.06\% & 29.75\% & 32.79\% & 30.04\% \\
\rowcolor{gray!10}
\shortstack[l]{\textit{Per report:} Mean Ques correct w/ report, incorrect w/o} 
& 8.93 & 11.88 & 10.42 & 11.90 & 13.12 & 12.02 \\
\shortstack[l]{\textit{Per report:} Std Ques correct w/ report, incorrect w/o} 
& 4.23 & 4.51 & 4.15 & 4.74 & 3.60 & 4.67 \\
\bottomrule
\end{tabular}
}
\caption{
Statistics of the generated MCQA dataset before filtering, evaluated using various metrics. After filtering, approximately 25\%, 27\%, and 31\% of questions requiring the report for correct answers were retained for CheXpertPlus\_MIMIC, CheXpertPlus\_Chex\_MIMIC, and MAIRA2, respectively. On average, 8 to 13 such questions per report were retained across both GT and GEN reports. 
}
\label{tab:filtering_results}
\end{table*}

\begin{table*}[t]
\centering
\scriptsize
\renewcommand{\arraystretch}{1.2}
\adjustbox{max width=\textwidth}{%
\begin{tabular}{l >{\ttfamily}r >{\ttfamily}r >{\ttfamily}r >{\ttfamily}r >{\ttfamily}r >{\ttfamily}r}
\toprule
\textbf{ID} & \multicolumn{3}{c}{\textbf{Questions from Ground-Truth Report}} & \multicolumn{3}{c}{\textbf{Questions from Generated Report}} \\
\cmidrule(lr){2-4} \cmidrule(lr){5-7}
& \makecell{CheXpertPlus\\MIMIC} & \makecell{CheXpertPlus\\CheX MIMIC} & MAIRA-2
& \makecell{CheXpertPlus\\MIMIC} & \makecell{CheXpertPlus\\CheX MIMIC} & MAIRA-2 \\
\midrule
0  & 23581 & 23753 & 23998 & 22206 & 21774 & 45365 \\
1  & 7227  & 7086  & 7224  & 589   & 639   & 2759  \\
2  & 2896  & 2855  & 2787  & 9140  & 15821 & 2801  \\
3  & 4775  & 4728  & 4739  & 1443  & 1582  & 4122  \\
4  & 6723  & 6771  & 6722  & 2589  & 2972  & 1181  \\
5  & 14375 & 14561 & 14449 & 8215  & 11966 & 3816  \\
6  & 7699  & 7647  & 7586  & 2676  & 2361  & 2391  \\
7  & 20562 & 20558 & 20591 & 16187 & 30311 & 44212 \\
8  & 8258  & 8374  & 8427  & 2530  & 1776  & 6323  \\
9  & 3383  & 3312  & 3367  & 798   & 976   & 1300  \\
10 & 19593 & 19872 & 19772 & 17326 & 22202 & 42859 \\
11 & 482   & 472   & 438   & 1068  & 443   & 386   \\
12 & 4679  & 4530  & 4616  & 1541  & 1321  & 2916  \\
13 & 3159  & 3131  & 3251  & 74    & 267   & 2393  \\
14 & 6184  & 6364  & 6264  & 4172  & 5860  & 1688  \\
15 & 7891  & 7988  & 7924  & 18546 & 8745  & 3319  \\
16 & 13452 & 13458 & 13335 & 4117  & 7917  & 4570  \\
17 & 2421  & 2229  & 2289  & 5479  & 1075  & 2339  \\
18 & 8244  & 8355  & 8296  & 4677  & 3903  & 5328  \\
19 & 10763 & 10854 & 10868 & 5266  & 5130  & 7467  \\
\bottomrule
\end{tabular}
}
\caption{Cluster-level question counts from ground-truth and generated reports across three models.}
\label{tab:ques_categorization_counts}
\end{table*}

\begin{figure*}[t]
\centering
\includegraphics[width=\textwidth]{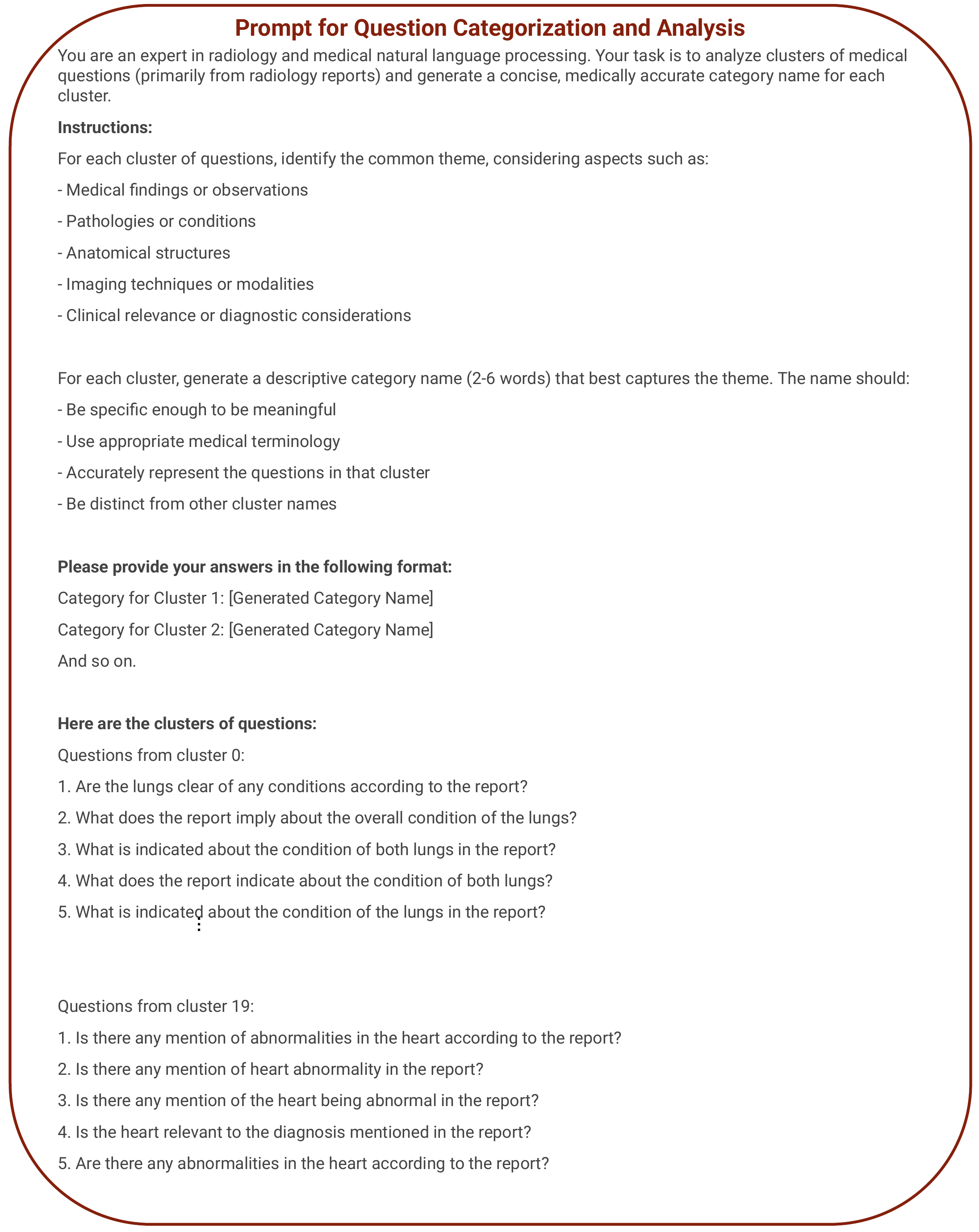}  
\caption{Prompts given to the language model for the Question Categorization and Analysis.}  
\label{fig:prompt_ques_categorization}  
\end{figure*}

{\normalsize
\begin{table*}[t]
\centering
\renewcommand{\arraystretch}{1.3}
\rowcolors{2}{gray!20}{white}
\adjustbox{max width=\textwidth}{%
\begin{tabular}{p{12cm} l}
\toprule
\textbf{Questions from MCQ Dataset Generation Module} & \textbf{Cluster Name} \\
\midrule
Are the lungs clear of any conditions according to the report? & Lung Condition Assessment \\
Are degenerative changes noted in any part of the spine other than the thoracic region? & Thoracic Spine Degeneration \\
Does the report mention the presence of fluid in the lungs or pleural space? & Pleural Fluid Presence \\
Is the right hemidiaphragm normally positioned or elevated according to the radiology report? & Diaphragm Position Abnormality \\
Is there any mention of pulmonary vascular congestion in the report? & Pulmonary Vascular Condition \\
Does the report mention the heart being normal in size? & Heart Size Normalcy \\
Are there any opacities mentioned in the radiology report and if so, where? & Lung Opacity and Infiltrates \\
Can pleural effusion be present according to the report? & Pleural Effusion Presence \\
Are there any other findings mentioned in the report? & Additional Radiologic Findings \\
Which condition is mentioned in the report regarding the aorta? & Aortic Condition \\
Does the report indicate pneumothorax as a finding? & Pneumothorax Diagnosis \\
Where are the surgical clips located that indicate a prior surgical procedure, according to the report? & Surgical Clip Location \\
What is the condition of the lower lobe of the right lung according to the report? & Lung Lobe Abnormality \\
Where are the calcified granulomas specifically located, as stated in the report? & Calcified Granuloma Location \\
Is there any mention of an abnormality in the cardiac and mediastinal contours? & Cardiac and Mediastinal Contours \\
Does the report mention the condition of the osseous structures? & Osseous Structure Condition \\
Is there any mention of a focal airspace opacity in the lungs? & Focal Airspace Opacity \\
Is there any mention of the lungs being clear and free of abnormalities? & Lung Clearness and Normalcy \\
What does the report specifically mention about lung volumes? & Lung Volume Condition \\
Is there any mention of abnormalities in the heart according to the report? & Heart Abnormality Presence \\
\bottomrule
\end{tabular}
}
\caption{Sample questions generated by the MCQ Dataset Generation module across different clinical categories. Each question corresponds to one of the 20 semantic clusters identified in Section~2.5.}
\label{tab:mcqa_cluster_examples}
\end{table*}
}




\end{appendices}


\bibliography{sn-bibliography}

\end{document}